\documentclass[lettersize,journal]{IEEEtran}
\usepackage{amsmath,amsfonts}
\usepackage{algorithmic}
\usepackage{algorithm}
\usepackage{array}
\usepackage{amsmath}
\usepackage{amssymb}
\usepackage[caption=true,font=footnotesize]{subfig}
\usepackage{textcomp}
\usepackage{stfloats}
\usepackage{url}
\usepackage[hidelinks]{hyperref} 
\usepackage{verbatim}
\usepackage{graphicx}
\usepackage{cite}
\usepackage{upgreek}
\usepackage{comment} 
\usepackage{multirow}
\usepackage{graphicx}
\usepackage[table,xcdraw]{xcolor}
 
\usepackage{bm} 
\begin{document}

\title{A Model Predictive Capture Point Control Framework for Robust Humanoid Balancing via Ankle, Hip, and Stepping Strategies}
\author{Myeong-Ju Kim$^{1}$, Daegyu Lim$^{1}$, Gyeongjae Park$^{1}$, {Kwanwoo Lee$^{1}$}, and Jaeheung Park$^{1,2}$,~\IEEEmembership{Member,~IEEE}
\thanks{This work was supported by the National Research Foundation of Korea (NRF) grant funded by the Korea government (MSIT) (No. 2021R1A2C3005914).}
\thanks{$^{1}$Myeong-Ju Kim, Daegyu Lim, Gyeongjae Park, Kwanwoo Lee and Jaeheung Park are with the Department of Intelligence and Information, Seoul National University, Republic of Korea. Contact : {\tt\small park73@snu.ac.kr}}%
\thanks{$^{2}$Jaeheung Park is also with the Advanced Institutes of Convergence Technology, Republic of Korea and with ASRI, RICS, Seoul National University, Republic of Korea.}
}

\markboth{Journal of \LaTeX\ Class Files,~Vol.~14, No.~8, August~2021}%
{Shell \MakeLowercase{\textit{et al.}}: A Sample Article Using IEEEtran.cls for IEEE Journals}

\IEEEpubid{0000--0000/00\$00.00~\copyright~2021 IEEE}

\maketitle

\begin{abstract}
The robust balancing capability of humanoids is essential for mobility in real environments. Many studies focus on implementing human-inspired ankle, hip, and stepping strategies to achieve human-level balance. In this paper, a robust balance control framework for humanoids is proposed. Firstly, a Model Predictive Control (MPC) framework is proposed for Capture Point (CP) tracking control, enabling the integration of ankle, hip, and stepping strategies within a single framework. Additionally, a variable weighting method is introduced that adjusts the weighting parameters of the Centroidal Angular Momentum damping control. Secondly, a hierarchical structure of the MPC and a stepping controller was proposed, allowing for the step time optimization. 
The robust balancing performance of the proposed method is validated through simulations and real robot experiments. Furthermore, a superior balancing performance is demonstrated compared to a state-of-the-art Quadratic Programming-based CP controller that employs the ankle, hip, and stepping strategies.
\end{abstract}


\section{Introduction}\label{Section/Introduction}

\IEEEPARstart{H}{umanoid} robots have been studied to achieve human-like walking capability in environments designed for human activities. However, their ability to adapt to disturbances caused by uneven terrain and potential collisions remains a challenge. To navigate these environments successfully, a robust balance control strategy that overcomes disturbances and ensures safe locomotion is essential.

{In order to obtain robust balancing performance in humanoid robots}, researchers have been inspired by human balance strategies, and several studies are conducted to explore them \cite{nashner1985organization, winter1995human, maki1997role}. The imitation and analysis of human balance strategies, including ankle, hip, and stepping strategies, have been explored using a simple model \cite{barin1989evaluation, kuo1993human, park2004postural}. The lessons learned from these investigations have been applied to humanoid robot research, with each strategy being implemented in a distinct manner. Specifically, the ankle strategy has been implemented through the Zero Moment Point (ZMP) control and the hip strategy has been implemented using Centroidal Angular Momentum (CAM) control. Lastly, the stepping strategy has been realized through the stepping control, which adjusts footstep position or step time.

\IEEEpubidadjcol In the early days of humanoid walking research, many studies were conducted to control ZMP, which is frequently referred to as an ankle strategy, by utilizing a simplified model. The Linear Inverted Pendulum Model (LIPM) \cite{kajita20013d, kajita2003biped} was introduced, which simplifies the dynamics of complex humanoid robots, and the linear relationship between the Center of Mass (CoM) and ZMP \cite{vukobratovic2004zero, popovic2005ground} is established. Many studies aimed to minimize the error between the reference ZMP and the actual ZMP, premised on the concept that if the ZMP is positioned within the support region, the robot will remain stable \cite{kajita2003preview, choi2006walking, kim2006experimental, kajita2010biped, joe2019robust}. Kajita et al. \cite{kajita2003preview} proposed a preview control method that considers a future reference ZMP trajectory, which compensates for ZMP errors by generating CoM trajectories. Choi et al. \cite{choi2006walking} calculates the desired CoM velocity to regulate ZMP and CoM errors, and utilizes the CoM Jacobian to generate the desired CoM velocity. Kim et al. \cite{kim2006experimental} proposed a state observer-based ZMP feedback controller that accounts for the joint elasticity of the robot. 
Kajita et al. \cite{kajita2010biped} calculates contact wrenches to track the reference ZMP and utilized an admittance control method to generate the calculated contact wrenches. Joe et al. \cite{joe2019robust} proposed a comprehensive ZMP control framework that combines the state feedback controller suggested in \cite{kim2006experimental} with the contact wrench controller proposed in \cite{kajita2010biped}.
While these studies have greatly contributed to the research on balancing in humanoid walking, the robot's ability to maintain balance relies only on ZMP control whose balancing capacity is limited by the size of the supporting polygon. Therefore, additional strategies are necessary to withstand strong disturbances.

Studies have been conducted to enhance the balancing capabilities of humanoid robots by exploring not only the ankle strategy utilizing ZMP control but also the hip strategy utilizing the robot's upper body. The hip strategy is commonly used in humanoids to control the CAM with the upper body. To account for the angular momentum that was disregarded in the LIPM, simple models such as Angular Momentum inducing inverted Pendulum Model (AMPM) or Linear Inverted Pendulum Plus Flywheel Model (LIPFM) have been proposed \cite{komura2005feedback, komura2005simulating, pratt2006capture}, and the dynamic relationship between the CoM and Centroidal Moment Pivot (CMP) has been defined \cite{pratt2006capture}. Based on this, many CAM control frameworks have been proposed to overcome external disturbances \cite{stephens2007humanoid, yi2016whole, wiedebach2016walking, schuller2021online, kim2022humanoid, ding2022dynamic}. 
Yi et al. \cite{yi2016whole} proposed a CAM control framework that generates a desired CAM through the hip joint and sequentially recovers the initial pose at a predetermined time when the CoM is perturbed by external disturbances. Schuller et al. \cite{schuller2021online} proposed a CAM control framework based on whole-body dynamics, in which the CAM tracking control and initial pose return control are operated by soft hierarchy-based Quadratic Programming (QP) optimization. In our previous CAM control approach \cite{kim2022humanoid}, we addressed the issue of degraded balancing performance caused by the conflict between initial pose control and CAM tracking control arising from the soft hierarchy in \cite{schuller2021online} by utilizing a control framework based on Hierarchical Quadratic Programming (HQP).
Ding et al. \cite{ding2022dynamic} proposed a CAM controller that plans arm trajectories to improve balancing performance using the Model Predictive Control (MPC) approach.
These methods have been shown to improve balancing performance. However, the amount of CAM generation in a robot is constrained by the joint limits of the robot and the self-collision avoidance, which in turn affects the robot's ability to maintain balance.

The stepping control, which adjusts the footstep position or step time in an adaptive manner to disturbances, has greatly improved the balancing performance of humanoid robots compared to the conventional walking control based on pre-determined footsteps and step time\cite{herdt2010online, khadiv2016step, jeong2017biped, joe2018balance, jeong2019robust1, khadiv2020walking, kim2023foot}. Under the notion that the position of the CoM converges to the Capture Point (CP) using LIPM dynamics, several stepping algorithms have been proposed \cite{khadiv2016step, jeong2017biped, jeong2019robust1,khadiv2020walking}. 
Khadiv et al. \cite{khadiv2016step, khadiv2020walking} proposed a framework that employs QP optimization based on the CP end-of-step dynamics to calculate the footstep position and step time. This approach tracks the pre-planned CP offset during walking by adjusting the footstep position and step time. Jeong et al. \cite{jeong2017biped, jeong2019robust1} proposed a stepping control method that also utilizes CP end-of-step dynamics, where the ankle torque is pre-calculated in response to external disturbances, and the step time and footstep position are optimized accordingly.
Furthermore, many control frameworks have been proposed for adjusting footstep position against disturbances by utilizing MPC optimization \cite{herdt2010online, joe2018balance, kim2023foot}. Herdt et al. \cite{herdt2010online} proposed an MPC framework that extends the framework proposed by Wieber et al. \cite{wieber2006trajectory}, by automatically adjusting the footstep position to control the CoM velocity and the ZMP error. Joe et al. \cite {joe2018balance} proposed a framework for adjusting footstep position when the desired ZMP generated to reduce ZMP error is limited by ZMP constraints. 
The continuous evolution of stepping algorithms has significantly contributed to humanoid balance control.  


With the advancement of each balance control strategy, many studies have been proposed with the aim of integrating these strategies to enhance balancing capability \cite{shafiee2017robust, ding2021versatile, ding2022robust, romualdi2022online}.
Shafiee-Ashtiani et al. \cite{shafiee2017robust} proposed a linear MPC framework that builds upon the MPC-based stepping controller suggested by \cite{herdt2010online} to combine the three balance strategies (ankle, hip, and stepping strategies).
In this approach, the ZMP and CoM velocity control problem proposed in \cite{herdt2010online} was changed to a ZMP and CP control problem. Furthermore, the control inputs were expanded to include the change of the centroidal moment, resulting in improved control performance.
Ding et al. \cite{ding2021versatile, ding2022robust} proposed a nonlinear MPC framework using a nonlinear Inverted Pendulum Flywheel (IPF) model. By considering the nonlinear relationship between CoM and ZMP in IPF as a quadratic constraint, the method adjusts the ZMP and body angles, as well as the footstep position to compensate for large disturbances.
Romualdi et al. \cite{romualdi2022online} proposed a nonlinear MPC framework for disturbance rejection based on centroidal dynamics, which involves the control of contact wrench, CAM, and footstep position. 
These studies developed a robust walking framework to cope with disturbances by integrating the three balance strategies through MPC. However, they did not consider step time adjustment algorithms, which play a crucial role in withstanding disturbances.

Studies that included the three balance strategies and a step time adjustment were also proposed \cite{aftab2012ankle, nazemi2017reactive, jeong2019robust}. 
Aftab et al. proposed an MPC-based balance control framework that integrates the three balance strategies and step time adjustment to address disturbances \cite{aftab2012ankle}. To avoid a biased output of the smallest step time during step time optimization, a swing foot acceleration cost was introduced to adjust the step time appropriately. However, due to the nonlinearity arising from step time optimization, the algorithm could not be used in real-time and was only applicable to standing situations, not during walking control.
Nazemi et al. proposed a reactive walking pattern generator based on hierarchical structure \cite{nazemi2017reactive}. In this approach, the stepping controller proposed by \cite{khadiv2016step, khadiv2020walking} is first used to determine the footstep position and step time for the stepping strategy. Subsequently, the CP trajectory and body angle were adjusted using MPC to track the reference ZMP trajectory determined by the pre-planned footstep position and step time.
Jeong et al. \cite{jeong2019robust} proposed a QP-based optimization method to achieve the three balance strategies with step time adjustment for controlling the CP end-of-step. In this method, to avoid the variable coupling during the linearization for QP, the ZMP control input for ankle strategy was pre-computed using the instantaneous CP control method \cite{englsberger2011bipedal}, rather than through optimization. Based on this pre-computed ZMP control input, the CAM control and stepping control are performed through QP optimization.  
In \cite{nazemi2017reactive,jeong2019robust}, the robust balancing performance was validated against disturbances through simulations or experiments. However, the three balance strategies cannot be computed from a single framework in \cite{nazemi2017reactive,jeong2019robust}. Additionally, these methods are unable to consider future states and constraints beyond the current walking step, which can affect the balancing performance. This limitation arises from the control framework relying on CP end-of-step dynamics, which restricts the prediction horizon to the current step duration. 

In this paper, a robust balance control framework is proposed to overcome disturbances through the integration of the three balance strategies and step time optimization. In contrast to \cite{nazemi2017reactive,jeong2019robust}, our research does not restrict the horizon time of MPC based on the current step duration. The proposed framework expands upon our prior works \cite{kim2022humanoid, kim2023foot} that were developed for the same goal, i.e., CP tracking control. We combined the hip strategy in \cite{kim2022humanoid} and the ankle and stepping strategies in \cite{kim2023foot} into a single MPC framework. 
With novel ideas, we addressed several technical problems that arose during the integration process, thereby enhancing the coherence between control hierarchies in the proposed control framework.
The primary contributions of this paper can be summarized as follows:

\begin{enumerate}
    \item An MPC framework that integrates three balance strategies for CP tracking control is proposed as an extension of our prior work \cite{kim2023foot}. Unlike \cite{kim2023foot}, this framework optimizes not only ZMP and stepping control but also CAM control. Additionally, it enhances CAM control through MPC, rather than using the heuristic CAM calculation method proposed in our previous study \cite{kim2022humanoid}. 
    \item A novel variable weighting method for CAM control is proposed. This method adjusts the weighting parameters of CAM damping control during the time horizon of the MPC to enhance the CP control performance.
    \item A hierarchical structure of the proposed MPC and stepping controller enables optimization of step time. Furthermore, an approach for determining stepping control parameters based on the MPC is suggested, which achieves better control performance than the previous parameter selection approach \cite{khadiv2016step, khadiv2020walking, kim2023foot}.
    \item { A novel HQP-based whole-body IK controller is proposed, focusing on nominal pose tracking and CAM control, while ensuring a smooth transition into initial pose recovery.}
    \item 
    The proposed method is validated through extensive simulations and real robot experiments, showing superior balancing performance compared to a state-of-the-art QP-based CP controller \cite{jeong2019robust} that incorporates the three balance strategies.
\end{enumerate}

This paper is organized as follows. Section \ref{Section/Fundamentals} introduces the LIPFM and CP dynamics, followed by an overview of the walking control framework in Section \ref{Section/Overall Framework}. The proposed MPC framework is detailed in Section \ref{Section/MPC Framework}, and the MPC-based stepping controller is described in Section \ref{Section/Stepping Controller}. {Section \ref{Section/WBIK} presents the HQP-based WBIK.} Simulation and experimental results validating the proposed algorithm are provided in Section \ref{Section/Results}, while {Section \ref{Section/Discussion} discusses the framework and implementation details.} Finally, the conclusion is presented in Section \ref{Section/Conclusion}. Henceforth, the proposed MPC framework is referred to as CP--MPC.

\section{Fundamentals}
\label{Section/Fundamentals}
\subsection{Linear Inverted Pendulum Plus Flywheel Model}

The LIPFM is a linear abstract model that was developed to address the CAM of humanoid robots \cite{pratt2006capture}.
The LIPFM assumes that the total mass of the robot is concentrated at the CoM, and the height of the CoM from the ground is considered to be constant. Unlike the LIPM \cite{kajita20013d, kajita2003biped}, the LIPFM is capable of handling reaction torque by means of the rotational motion of a flywheel located at the CoM. 
The dynamic equation of the LIPFM is expressed in terms of the relationship between the CoM and the CMP, and is defined as follows,
\begin{equation}
{\Ddot{c}_x}={\omega^2}(c_x-p_{x}),
\label{eq/COM-CMP}
\end{equation}
\begin{equation}
\label{eq/CMP-ZMP-Moment}
{p_{x}} = {z_{x}} + \frac{{\tau_y}}{mg},
\end{equation}
where $c_x$, $p_{x}$, and $z_{x}$ denote the positions of CoM, CMP, and ZMP in the x-direction, respectively. $\omega=\sqrt{g/c_z}$ is the natural frequency, ${g}$ is the gravitational acceleration, and $c_z$ is the height of the CoM from the ground. $\tau_y$ represents the reaction torque of the flywheel in the y-direction. 
A detailed derivation of LIPFM dynamics is presented in \cite{pratt2006capture}.

The dynamics of LIPFM in the y-direction can be derived in the same way as the x-direction, and the dynamics in each direction can be dealt with independently. 
Therefore, in this study, the dynamics were derived only in the x-direction for conciseness of the paper, except for Section \ref{Section/Stepping Controller}, where the step time variable couples both x- and y-direction variables.

\subsection{Capture Point dynamics based on LIPFM}
\label{Section/Fundamentals/CP dynamics}
This section provides a brief introduction to the CP dynamics based on LIPFM.
The concept of CP, introduced in \cite{pratt2006capture, hof2008extrapolated}, has been extensively utilized in various studies as a control variable for stabilizing the CoM of a robot and as an indicator of its balance state \cite{englsberger2011bipedal, englsberger2016combining, jeong2017biped, joe2018balance, morisawa2012balance, wiedebach2016walking, griffin2017walking, joe2019robust}.
The dynamics of the CP is expressed as a linear combination of horizontal position and velocity of the CoM as below,
\begin{equation}
\xi_x = c_x + \frac{{\Dot{c}_x}}{\omega}, 
\label{eq/CP-COM}
\end{equation}
where $\xi_x$ represents the CP in the x-direction. 
The LIPFM also allows the CP to be expressed as a dynamic relationship with the CMP. By combining (\ref{eq/COM-CMP}) with the time derivative of (\ref{eq/CP-COM}), the dynamics of the CP--CMP can be derived as   
\begin{equation} 
\Dot{\xi}_x = \omega(\xi_x-p_{x})=\omega(\xi_x-({z_{x}} + \frac{{\tau_y}}{mg})).
\label{eq/CP-CMP}
\end{equation}
Assuming the constant CMP $p_x$ within a step duration $T$, and considering the time elapsed after the start of the swing phase as $t$, the behavior of the CP can be defined as follows,
\begin{equation}
{\xi}_{x,T} = (\xi_{x}-p_x)e^{\omega (T-t)} + p_x.
\label{eq/CP eos}
\end{equation}
Equation (\ref{eq/CP eos}), known as the CP end-of-step dynamics \cite{englsberger2011bipedal, jeong2019robust, jeong2019robust1}, provides a means to predict the CP at the end of a current step by using the current CP, current time, and CMP as inputs.

In Section \ref{Section/MPC Framework}, the CP--CMP dynamics (\ref{eq/CP-CMP}) is employed as a prediction model in the proposed CP--MPC. Additionally, the CP end-of-step dynamics (\ref{eq/CP eos}) is utilized for the stepping controller presented in Section \ref{Section/Stepping Controller}.

\section{Overall Walking Control Structure}
\label{Section/Overall Framework}
\begin{figure*}[t]
      \centering
      \includegraphics[width=0.9\linewidth]{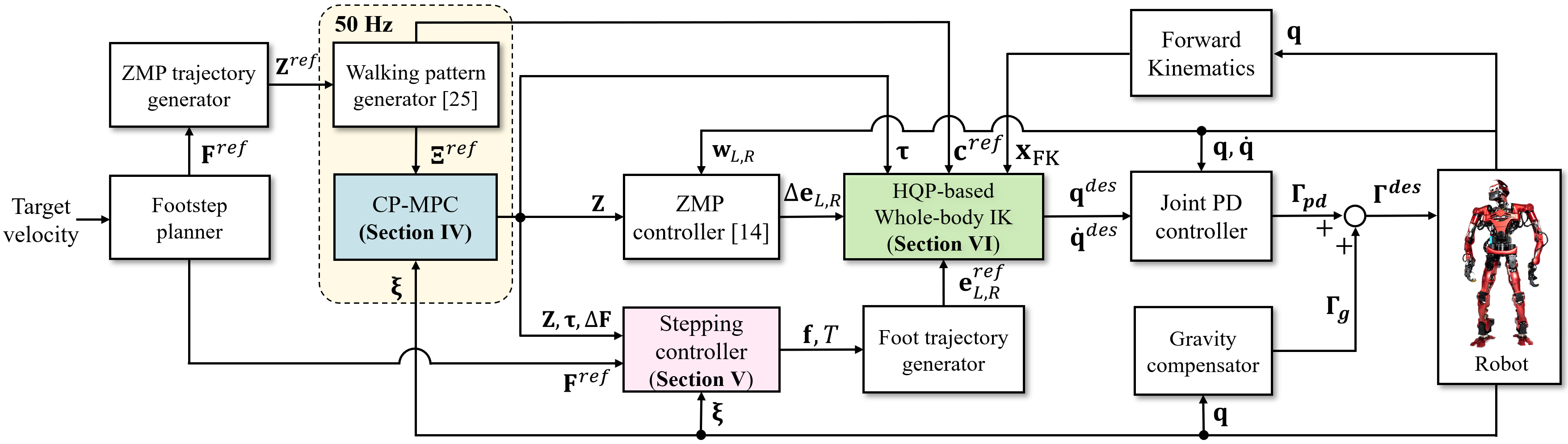} 
      \caption{{Overall walking control framework: Three balance strategies for CP tracking control via CP--MPC, footstep position and step time determination by the stepping controller, and implementation through HQP--WBIK.}}
      \label{figure/overallframe}
\end{figure*} 

This section provides an overview of the proposed walking control framework. The schematic diagram of the overall walking control framework is illustrated in Fig. \ref{figure/overallframe}. First, the footstep planner determines the positions of the footsteps $\mathbf{F}^{ref}=[\mathbf{F}_x^{ref}\,\mathbf{F}_y^{ref}]$ based on the target velocity command. Using the footstep positions as input, the ZMP trajectory generator creates reference ZMP trajectories $\mathbf{Z}^{ref}=[\mathbf{Z}_x^{ref}\,\mathbf{Z}_y^{ref}]\in\mathbb{R}^{N\times2}$. The walking pattern generator \cite{herdt2010online} then takes these reference ZMP trajectories as input and produces reference CP trajectories $\mathbf{\Xi}^{ref}=[\mathbf{\Xi}^{ref}_x\,\mathbf{\Xi}^{ref}_y]\in\mathbb{R}^{N\times2}$. {Note that our walking pattern generator does not include the footstep adjustment algorithm from \cite{herdt2010online}.}

The main goal of CP--MPC is to track the reference CP trajectories.  
The CP--MPC receives the reference CP trajectories as its control target and generates three optimized MPC variables, including ZMP control inputs $\mathbf{Z}=[\mathbf{Z}_x\,\mathbf{Z}_y]\in\mathbb{R}^{N\times2}$, centroidal moment control inputs $\bm{\uptau}=[\bm{\uptau}_y\,-\bm{\uptau}_x]\in\mathbb{R}^{N\times2}$, and footstep adjustments $\Delta \mathbf{F}=[\Delta\mathbf{F}_{x}\,\Delta\mathbf{F}_{y}]\in\mathbb{R}^{m\times2}$ from $\mathbf{F}^{ref}$ to track these trajectories. 
$N$ represents the prediction horizon for the CP--MPC, while $m$ denotes the number of pre-designed next footsteps within the prediction horizon $N$. 

The ZMP controller \cite{kajita2010biped} computes desired contact wrenches to achieve the desired ZMP $\mathbf{z}^{des}=[z_{x}^{des}\,z_{y}^{des}]\in\mathbb{R}^{2}$, which is the first element of the ZMP control inputs $\mathbf{Z}$. 
{Subsequently, the admittance controller compares the measured contact wrench $\mathbf{w}_{L,R}\in\mathbb{R}^{6}$ with the desired contact wrench to determine the changes in the position and orientation of the support foot with respect to the pelvis $\Delta \mathbf{e}_{L,R}\in\mathbb{R}^{6}$, where both wrenches reflect only the z-direction position and roll/pitch  orientation components.}
The subscript $L$ and $R$ signify the left and right, respectively.

The stepping controller calculates the next footstep position $\mathbf{f}=[{f}_{x}\,{f}_{y}]\in\mathbb{R}^{2}$ and the step time $T$ to achieve the footstep adjustment $\Delta \mathbf{f}_1=[\Delta {f}_{x,1}\, \Delta{f}_{y,1}]\in\mathbb{R}^{2}$ planned by CP--MPC. 
Specifically, the controller primarily adjusts the step time $T$ based on CP end-of-step dynamics (\ref{eq/CP eos}) to ensure that the $\mathbf{f}$ achieves $\Delta \mathbf{f}_1$ as closely as possible.
Here, $\Delta\mathbf{f}_1$ refers to the first element of $\Delta \mathbf{F}$, representing the adjustment of the next footstep. 
After determining $\mathbf{f}$ and $T$, the foot trajectory generator creates reference foot trajectories $\mathbf{e}^{ref}_{L,R}\in\mathbb{R}^{6}$.  
In addition, when a step change occurs, the footstep planner generates footsteps based on the target velocity command relative to the supporting foot.

{
The desired centroidal moment $\bm{\tau}^{des}=[\tau_{y}^{des},-\tau_{x}^{des}]\in\mathbb{R}^{2}$, which is the first element of the centroidal moment control input $\bm{\uptau}$, is integrated to obtain the desired CAM. In this study, only the CAM in the x- and y-directions is computed from the CP--MPC based on CP--CMP dynamics (\ref{eq/CP-CMP}), while the CAM in the z-direction is controlled to remain zero to compensate for momentum induced by the swing foot motion and to prevent yaw slip at the support foot.

{In the final stage, the HQP-based whole-body IK receives feedback from the FK results, represented as $\mathbf{X}_{FK}=[\mathbf{c}~\mathbf{e}~\mathbf{x}_{pelv}~\mathbf{x}_{chest}~\mathbf{x}_{hand}]^T$, and determines the desired joint angles $\mathbf{q}^{des}$ and joint velocities $\mathbf{\dot{q}}^{des}$ required to achieve the reference CoM trajectory $\mathbf{c}^{ref}\in\mathbb{R}^{3}$, reference foot trajectories $\mathbf{e}^{ref}_{L,R}\in\mathbb{R}^{6}$, support foot pose displacement $\Delta \mathbf{e}_{L,R}\in\mathbb{R}^{6}$, and desired CAM $\mathbf{h}^{des}\in\mathbb{R}^3$.
Then, a joint PD controller computes the torques $\bm{\Gamma}_{pd}$ to track the desired joint trajectory, applied alongside the gravity compensation torque $\bm{\Gamma}_{g}$ computed using the contact-consistent control framework \cite{park2006contact}.}}

\section{CP--MPC: Model Predictive Control Framework for Capture Point Tracking}
\label{Section/MPC Framework}
In this section, we extend our previous framework \cite{kim2023foot} to propose CP--MPC, an MPC framework integrating ankle, hip, and stepping strategies for balance control.
CP--MPC leverages the ZMP, footstep position, and centroidal moment to enhance CP control under strong external disturbances. The following subsections provide a detailed description of the proposed framework.

\subsection{Prediction of Future Trajectory}
\label{Section/MPC Framework/Prediction model} 
In order to formulate the CP--MPC, the CP--CMP dynamics described in (\ref{eq/CP-CMP}) is utilized as a prediction model. Equation (\ref{eq/CP-CMP}) can be discretized with the piecewise ZMP, centroidal moment, and sampling time $T_s$ as below,
\begin{equation} 
\label{eq/discrete CP-CMP}
\xi_{x,k+1} = A\xi_{x,k} + \mathbf{B}\left[z_{x,k}\quad\tau_{y,k}\right]^T
\end{equation}
where $A = e^{\omega T_s}, \mathbf{B} = \left[1-e^{\omega T_s}\quad \frac{1-e^{\omega T_s}}{mg}\right]$.
The CMP $p_{x}$ can be decomposed into ZMP $z_{x}$ and centroidal moment $\tau_{y}$ as shown in (\ref{eq/CP-CMP}), and two variables are independently treated as control inputs of the CP--MPC.
By recursive application of (\ref{eq/discrete CP-CMP}), the predicted future trajectories of CP over the time horizon $N$ that emanate from the current CP $\xi_{x,k}$ can be expressed as below,
\begin{equation}
\label{eq/reculsive dynamics}
\mathbf{\Xi}_x=\bm{\Phi}_\xi \xi_{x,k} + \bm{\Phi}_p\mathbf{P}_x ,
\end{equation}
\begin{equation}
\bm{\Phi}_\xi=
\begin{bmatrix}
A \\ \vdots \\ A^N 
\end{bmatrix}
,\quad
\bm{\Phi}_p=
\begin{bmatrix}
A^{0}B & \cdots & 0 \\ \vdots & \ddots & \vdots \\ A^{N-1}B & \cdots & A^{0}B 
\end{bmatrix}
,
\end{equation}
\begin{equation} 
\mathbf{\Xi}_x = 
\begin{bmatrix}
\xi_{x,k+1} \\ \vdots \\ \xi_{x,k+N} 
\end{bmatrix}
,\quad
\mathbf{P}_x = 
\begin{bmatrix}
z_{x,k} \\ \tau_{y,k} \\ \vdots \\ z_{x,k+N-1} \\ \tau_{y,k+N-1}
\end{bmatrix}\label{eq/reculsive dynamics2}
.
\end{equation}
$\bm{\Xi}_x \in \mathbb{R}^{N}$ represents the predicted future CP trajectory in the x-direction and $\mathbf{P}_x \in \mathbb{R}^{2N}$ denotes the future inputs of the ZMP and centroidal moment. The matrix $\bm{\Phi}_{\xi}\in \mathbb{R}^{N}$ defines the dynamic relationship between the current CP $\xi_{x,k}$ and future CP trajectory $\bm{\Xi}_x$. The matrix $\bm{\Phi}_{p}\in \mathbb{R}^{N\times 2N}$ defines the relationship between future inputs $\mathbf{P}_{x}$ and future CP trajectory $\bm{\Xi}_x$. 
{In $\mathbf{P}_{x}$, the elements of ZMP $z_x$ and centroidal moment $\tau_y$ are sequentially incorporated. For intuitive representation, we separately define the series of ZMP vectors as $\mathbf{Z}_x=[z_{x,k}\,z_{x,k+1}\cdots z_{x,k+N-1}]^T\in \mathbb{R}^{N}$ and the series of centroidal moment vectors as $\bm{\uptau}_y=[\tau_{y,k}\,\tau_{y,k+1}\cdots\tau_{y,k+N-1}]^T\in \mathbb{R}^{N}$.}   
 
\subsection{Problem Setup for MPC Optimization}
\label{Section/MPC Framework/Cost function}
This section presents the problem formulation for MPC optimization. 
The cost function and constraints of CP--MPC are formulated {using (\ref{eq/reculsive dynamics})-(\ref{eq/reculsive dynamics2})} and are as follows,
\begin{align}
\label{eq/CP-MPC cost function}
& \underset{\mathbf{P}_{x},\Delta \mathbf{F}_x}{\text{min}} & & 
\|\mathbf{\Xi}_{x}-\mathbf{\Xi}^{ref}_{x}\|^2_{\mathbf{w}_{\xi}} + 
\|\boldsymbol{\uptau}_{y} + \mathbf{K}_d \mathbf{h}_y\|^2_{\mathbf{w}_{\tau}}  \\
& & +&\|\Delta\mathbf{F}_{x}\|^2_{\mathbf{w}_{F}} + \|\Delta\mathbf{P}_{x}\|^2_{\mathbf{w}_{p}} \nonumber\\
& \text{s. t.} & & \underline{\mathbf{Z}}_x\leq \begin{bmatrix}
\mathbf{I}_N & -\mathbf{S}
\end{bmatrix}
\begin{bmatrix} 
\mathbf{Z}_x\\ \Delta\mathbf{F}_x \label{eq/CP-MPC constraint}
\end{bmatrix}
 \leq \overline{\mathbf{Z}}_x\\
& & &\underline{\boldsymbol{\uptau}}_y\leq \boldsymbol{\uptau}_y \leq \overline{\boldsymbol{\uptau}}_y \nonumber\\
& & & \Delta\underline{\mathbf{F}}_{x}\leq \Delta\mathbf{F}_x \leq \Delta\overline{\mathbf{F}}_{x} \nonumber \\
& & & {\xi^{ref}_{x,k+N}=\xi_{x,k+N}.} \nonumber 
\end{align}

The cost function in (\ref{eq/CP-MPC cost function}) is composed of four cost terms in total. The first cost term functions as a reference CP trajectory tracking control and is weighted by a positive diagonal weighting matrix $\mathbf{w}_{\xi}\in\mathbb{R}^{N\times N}$. The second cost term plans the {centroidal moment $\boldsymbol{\uptau}_y$ that drives the CAM $\mathbf{h}_y$} to zero when the disturbance is small. Here, $\mathbf{K}_d\in\mathbb{R}^{N\times N}$ is a damping matrix which is positive and diagonal and $\mathbf{h}_y\in\mathbb{R}^{N}$ is the CAM in y-direction which is an integration of {centroidal moment} $\boldsymbol{\uptau}_y$. This term is weighted by a positive diagonal matrix, $\mathbf{w}_{\tau}\in\mathbb{R}^{N\times N}$. The diagonal terms of $\mathbf{w}_{\tau}$ are adjusted based on the magnitude of the ZMP control inputs to improve the CP control performance. This adjustment algorithm is explained in Section \ref{Section/Variable Weighting}. The third cost term regulates the additional footstep adjustments from the pre-planned footsteps. This term is weighted by a positive diagonal matrix $\mathbf{w}_F\in\mathbb{R}^{m\times m}$. The last cost term regulates the change of the control input to prevent rapid changes and generate a smooth control input. This term is weighted by positive diagonal matrix $\mathbf{w}_{p}\in\mathbb{R}^{2N\times 2N}$. The MPC variables consist of $\mathbf{P}_{x}$ and $\Delta\mathbf{F}_x$, where {$\mathbf{P}_{x}\in\mathbb{R}^{2N}$ represents the sequence of ZMP $z_{x,k}$ and centroidal moment $\tau_{y,k}$, treated as control inputs}, while $\Delta\mathbf{F}_x=[\Delta{f}_{x,1} \Delta{f}_{x,2} \cdots \Delta{f}_{x,m}]^{\textrm{T}}\in\mathbb{R}^m$ represents a vector consisting of additional footstep adjustments in the x-direction. {Here, although $\Delta\mathbf{F}_x$ is not defined as a variable in CP--CMP dynamics (\ref{eq/discrete CP-CMP}), it is utilized as an optimization variable to relax the ZMP constraint.}

{The constraints in (\ref{eq/CP-MPC constraint}) consist of three inequality conditions and one equality condition.}
The first constraint refers to the ZMP constraint, which confines the ZMP control input $\mathbf{Z}_x\in\mathbb{R}^{N}$ within the support polygon. The vectors $\overline{\mathbf{Z}}_x\in\mathbb{R}^{N}$ and $\underline{\mathbf{Z}}_x\in\mathbb{R}^{N}$ indicate the upper and lower limits of ZMP inputs, respectively.
{These limits are approximated by modeling the foot as a rectangle, defined by the foot's width and length relative to the reference ZMP $\mathbf{Z}^{ref}$, while taking yaw rotation into account.}
{To prevent the desired ZMP from being generated outside the support polygon due to approximation errors, we approximate the rectangular foot size such that the approximated support area falls within a range smaller than the actual foot shape-based support area.}
The matrix $\mathbf{I}_N\in\mathbb{R}^{N\times N}$ refers to an $N$ size identity matrix and the matrix $\mathbf{S}\in\mathbb{R}^{N\times m}$ represents a selection matrix whose element ${s}_{k,m}$ is 1 or 0. 
{The element ${s}_{k,m}$ is set to 1 to activate $\Delta{f}_{x,m}$ when time step $k$ is included in the $m$-th footstep after the current Single Support Phase (SSP), and it is set to 0 when time step $k$ is not included in the $m$-th footstep, including the current SSP.}
Using $\Delta\mathbf{F}_x$ to relax the ZMP constraint in situations where ZMP control inputs are limited improves the performance of CP control by increasing flexibility in generating ZMP control inputs.
{The second constraint refers to the limit on the amount of centroidal moment that can be produced by the robot's motion. The vectors $\overline{\boldsymbol{\uptau}}_y\in\mathbb{R}^{N}$ and $\underline{\boldsymbol{\uptau}}_y\in\mathbb{R}^{N}$ indicate the upper and lower bounds of the centroidal moment, respectively. These constraints were experimentally determined by considering joint position and velocity limits.} The third constraint denotes the kinematic constraint of the additional footsteps to prevent singularities and self-collisions {between legs}. The vectors $\Delta\overline{\mathbf{F}}_x\in\mathbb{R}^{m}$ and $\Delta\underline{\mathbf{F}}_x\in\mathbb{R}^{m}$ refer to the upper and lower bounds of additional footsteps adjustment. {The last constraint signifies the terminal constraint imposed to ensure the convergence of CP tracking.}

{The number of footsteps $m$ in the optimization is set sufficiently large to accommodate the minimum walking duration and the time horizon. If the footsteps in the time horizon is less than $m$, the unnecessary footstep constraints are dynamically excluded from MPC by setting zero for the elements in $\mathbf{S}$ and constraining the footstep adjustment.}

{In (\ref{eq/CP-MPC cost function}), the generation of $\boldsymbol{\uptau}_y$ and $\Delta\mathbf{F}_x$ is regulated by the second and third cost terms, respectively. Therefore, for small disturbances, CP control is primarily achieved through ZMP control. For large disturbances, the variable weighting approach in Section \ref{Section/Variable Weighting} facilitates the generation of $\boldsymbol{\uptau}_y$ to reduce CP errors. Additionally, $\Delta\mathbf{F}_x$ is generated to relax the ZMP constraints of the next footsteps, leading to larger ZMP inputs and a reduction in CP errors. The weighting ratio of $\mathbf{w}_{\tau}$ and $\mathbf{w}_{F}$ can determine the preferred balance strategy (hip or stepping strategy) in the presence of large disturbances.}
The values of weighting parameters and constraints used in the experiments are summarized in Table \ref{table/Parameters}.

\begin{table}[t]
\caption{{Parameters used in the controllers for simulations and real robot experiments.}}
\label{table/Parameters}
\resizebox{\columnwidth}{!}{%
\begin{tabular}{|lll|}
\hline

\multicolumn{3}{|c|}{\cellcolor[HTML]{c6eefe}CP--MPC}                \\ \hline
\multicolumn{1}{|c|}{Parameters} & \multicolumn{2}{c|}{Value}                   \\ \hline
\multicolumn{1}{|l|}{$\mathbf{w}_{\xi,x,y}$}        & \multicolumn{2}{l|}{\begin{tabular}[c]{@{}l@{}}$10.0\,(i=1), 5.0\,(i=2 \sim N-10),$\\ $100.0\,(i = N-10 \sim N)$\end{tabular}} \\ \hline
\multicolumn{1}{|l|}{$\mathbf{w}_{p,x,y}$}        & \multicolumn{2}{l|}{\begin{tabular}[c]{@{}l@{}}$0.1\,(i=1), 10.0\,(i=2 \sim N-10),$\\ $0.1\,(i = N-10 \sim N)$\end{tabular}} \\ \hline
\multicolumn{1}{|l|}{$\mathbf{w}_{F,x,y}$}        & \multicolumn{2}{l|}{$0.001$} \\ \hline
\multicolumn{1}{|l|}{$\mathbf{w}_{\tau,y}$}        & \multicolumn{2}{l|}{$1.0\times10^{-6} \,(\Delta z_{min,x} : 0.05) \xrightarrow{} 0 \,(\Delta z_{max,x} : 0.1)$} \\ \hline
\multicolumn{1}{|l|}{$\mathbf{w}_{\tau,x}$}        & \multicolumn{2}{l|}{$1.0\times10^{-6} \,(\Delta z_{min,y} : 0.04) \xrightarrow{} 0 \,(\Delta z_{max,y} : 0.07)$} \\ \hline
\multicolumn{1}{|l|}{${\mathbf{K}}_{d}$} & \multicolumn{2}{l|}{$50.0\,\mathbf{I}_N$} \\ \hline
\multicolumn{1}{|l|}{$\underline{\mathbf{Z}}_{x},\overline{\mathbf{Z}}_{x}\,[m]$}        & \multicolumn{2}{l|}{$(\mathbf{Z}_x^{ref}-0.09,\,\mathbf{Z}_x^{ref}+0.12)$} \\ \hline
\multicolumn{1}{|l|}{$\underline{\mathbf{Z}}_{y},\overline{\mathbf{Z}}_{y}\,[m]$}        & \multicolumn{2}{l|}{$(\mathbf{Z}_y^{ref}-0.07,\,\mathbf{Z}_y^{ref}+0.07)$} \\ \hline
\multicolumn{1}{|l|}{$\Delta\underline{\mathbf{F}}_x,\Delta\overline{\mathbf{F}}_x\,[m]$}        & \multicolumn{2}{l|}{$(-0.2,0.2)$ w.r.t$\;$support foot} \\ \hline
\multicolumn{1}{|l|}{$\Delta\underline{\mathbf{F}}_y,\Delta\overline{\mathbf{F}}_y\,[m]$}        & \multicolumn{2}{l|}{$(-0.1,0.03)$/$(-0.03,0.1)$ w.r.t$\,$ left/right support foot} \\ \hline
\multicolumn{1}{|l|}{$\underline{\boldsymbol{\uptau}}_{y,x},\overline{\boldsymbol{\uptau}}_{y,x}\,[N\cdot m]$} & \multicolumn{2}{l|}{$(-15.0,15.0)$ (Simulation) /$\,(-7.0,7.0)$ (Experiment)} \\ \hline

\multicolumn{3}{|c|}{\cellcolor[HTML]{FFE6FC}Stepping controller}  \\ \hline
\multicolumn{1}{|c|}{Parameters} & \multicolumn{2}{c|}{Value}                   \\ \hline
\multicolumn{1}{|l|}{$\mathbf{w}_{f,x,y}$}        & \multicolumn{2}{l|}{$1000.0$}                    \\ \hline
\multicolumn{1}{|l|}{$\mathbf{w}_{b,x,y}$}        & \multicolumn{2}{l|}{$3000.0$}                    \\ \hline
\multicolumn{1}{|l|}{$w_\gamma$}     & \multicolumn{2}{l|}{$1.0$}                 \\ \hline
\multicolumn{1}{|l|}{$\underline{{f}}_{x,y},\overline{{f}}_{x,y}\,[m]$}         & \multicolumn{2}{l|}{$({f}_{nom,x,y}-0.05,{f}_{nom,x,y}+0.05$)}                \\ \hline
\multicolumn{1}{|l|}{$\underline{{b}}_{x,y},\overline{{b}}_{x,y}\,[m]$}         & \multicolumn{2}{l|}{$({b}_{nom,x,y}-0.10,{b}_{nom,x,y}+0.10)$}                \\ \hline 
\multicolumn{1}{|l|}{$\underline{T},\overline{T}\,[s]$}  & \multicolumn{2}{l|}{$(T_{nom}-0.2,T_{nom}+0.2)\;$ /$\;\gamma = e^{\omega T}$}                \\ \hline

\multicolumn{3}{|c|}{\cellcolor[HTML]{C6FDC6}HQP--WBIK}                \\ \hline
\multicolumn{1}{|c|}{Parameters}              & \multicolumn{2}{c|}{Value}                   \\ \hline
\multicolumn{1}{|l|}{$\mathbf{w}_{J}$}        & \multicolumn{2}{l|}{\begin{tabular}[c]{@{}l@{}}$120.0$ \text{(Foot pose)}, $10.0$ \text{(CoM)}, $0.1$ \text{(Hand pose)}, \\ $1.0$ \text{(Chest orientation)}, $5.0$ \text{(Pelvis orientation)}\end{tabular}}                    \\ \hline
\multicolumn{1}{|l|}{$\mathbf{w}_{A}$}        & \multicolumn{2}{l|}{$1.0$}                    \\ \hline
\multicolumn{1}{|l|}{$\mathbf{w}_{\dot{q}}$}  & \multicolumn{2}{l|}{$0.0001$} \\ \hline
\multicolumn{1}{|l|}{$\mathbf{K}_{p}$}         & \multicolumn{2}{l|}{\begin{tabular}[c]{@{}l@{}}$30.0$ \text{(Foot pose)}, $30.0$ \text{(CoM)}, $10.0$ \text{(Hand pose)}, \\ $10.0$ \text{(Chest orientation)}, $10.0$ \text{(Pelvis orientation)}\end{tabular}}                     \\ \hline
\multicolumn{1}{|l|}{$\mathbf{K}_{q}$}        & \multicolumn{2}{l|}{$50.0$}                    \\ \hline
\end{tabular}%
}
\end{table}

\subsection{Variable Weighting Parameter for CAM control}
\label{Section/Variable Weighting}
The CAM control approaches have been widely utilized in many studies to overcome large disturbances that affect robots. In general, a desired CAM of humanoids is initially generated to overcome external disturbances, but it {should converge} to zero in steady-state. 
This is because there is no need to generate constant CAM for the balancing purpose in steady-state, and furthermore, a constant desired CAM cannot be generated continuously due to the joint limits of the robot.
The convergence process of desired CAM can be implemented heuristically during a specific time period \cite{yi2016whole} or under certain conditions \cite{kim2022humanoid, schuller2021online, kanamiya2010ankle}. However, this issue is more commonly addressed through optimization methods by implementing a CAM regulation term or CAM damping control term in the cost function\cite{jeong2019robust, khazoom2022humanoid, romualdi2022online, ding2021versatile, park2021whole}.
In our case, the convergence process of desired CAM is also addressed through the CAM damping control term in MPC optimization. However, these approaches commonly suffer from a trade-off between generating the desired CAM to overcome the disturbance and the term that drives the CAM to zero, which can have a negative impact on the balancing performance.  

\begin{figure}[t]
      \centering
      \includegraphics[width=1.0\linewidth]{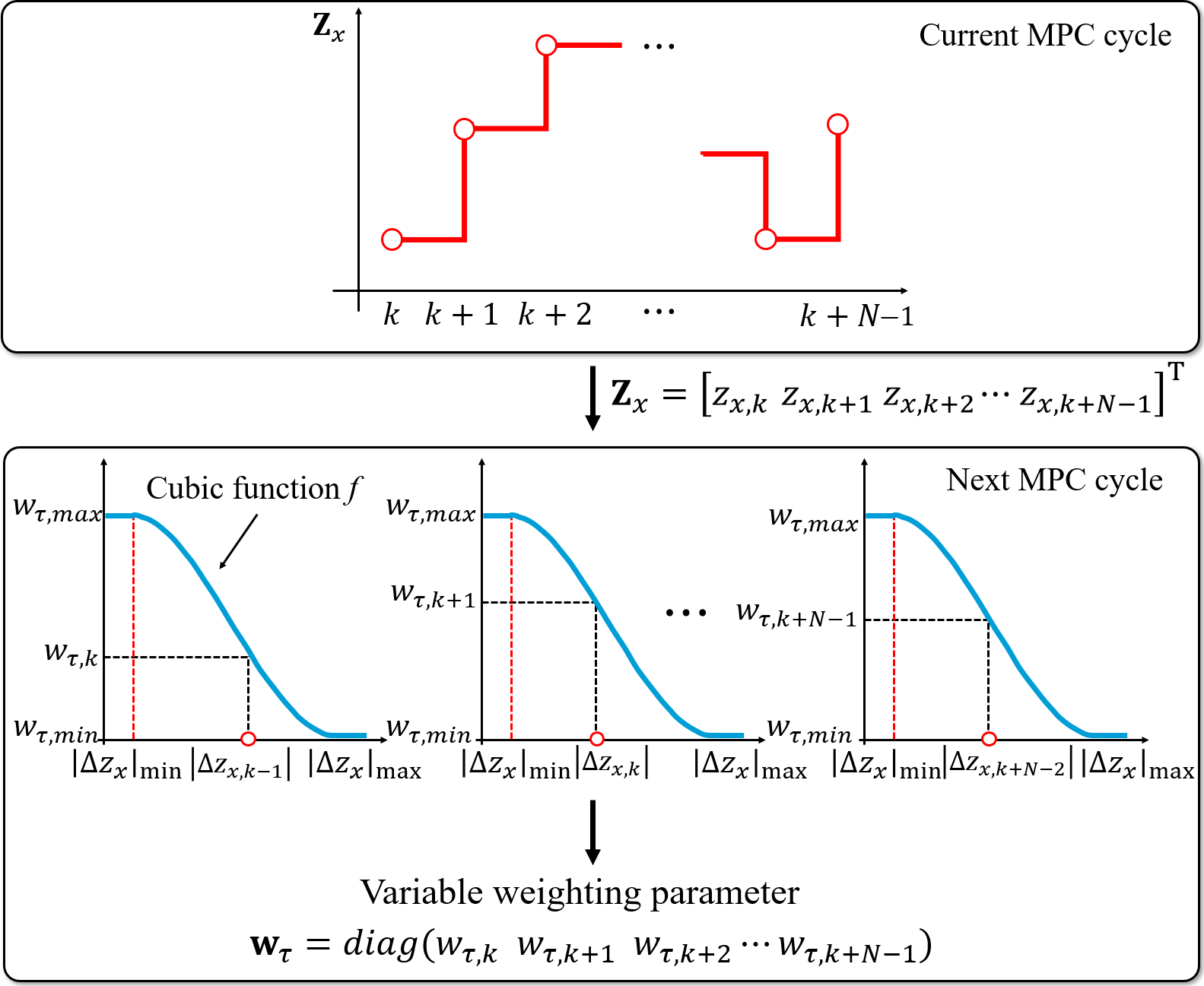} 
      \caption{{Decision process of variable weighting parameter.}}
      \label{figure/VariableWeighting}
\end{figure} 
To address this problem, a novel variable weighting approach based on MPC is proposed. 
The purpose of this approach is not to compromise the CP control performance when the robot is subjected to large disturbances, by reducing the weighting parameter $\mathbf{w}_{\tau}$ to reduce the influence of the damping cost term in (\ref{eq/CP-MPC cost function}).
This allows the robot to generate more control input $\bm{\uptau}$ in the presence of disturbances, thereby improving balance performance. Furthermore, in the steady-state when the disturbances diminish, the weighting parameter can be increased to dampen the generation of CAM.

Fig. \ref{figure/VariableWeighting} presents the scheme of the decision process of the variable weighting parameters in the x-direction.
In Fig. \ref{figure/VariableWeighting}, when a disturbance is applied to the robot, ZMP control inputs $\mathbf{Z}_x$ increase in CP--MPC to control the perturbed CP.
{Afterward, the weighting parameters $\mathbf{w}_{\tau}$ for the next MPC cycle are determined through a one-to-one mapping using a cubic function based on the magnitude of the delta ZMP inputs $|\Delta \mathbf{Z}_x|=|\mathbf{Z}_{x}-\mathbf{Z}_x^{ref}|\in\mathbb{R}^{N}$ in the current MPC cycle.}
{Note that the coefficients of the cubic function $f$ are determined by the following four conditions: $f(|\Delta z_x|_{max})=w_{\tau,min},\;f(|\Delta z_x|_{min})=w_{\tau,max},\;f'(|\Delta z_x|_{max})=f'(|\Delta z_x|_{min})=0$,} and $\mathbf{Z}_x^{ref}\in\mathbb{R}^{N}$ represents the reference ZMP trajectory in the x-direction based on the pre-planned footstep.
As a result, the damping cost term in (\ref{eq/CP-MPC cost function}) becomes less influential, and the centroidal moment $\bm{\uptau}_y\in\mathbb{R}^{N}$ is primarily generated to balance the robot against disturbances. 
The operational principle of this algorithm is based on the observation that the increased ZMP control inputs $\mathbf{Z}_x$ indicate that the robot is subjected to a substantial disturbance, thereby requiring more control inputs to accurately track the CP trajectory.
Therefore, the capability of CP control can be increased by increasing the generation of another control input $\bm{\uptau}_y$. The same process is applied in the y-direction, independent of the x-direction.

{Finally, the desired centroidal moment $\bm{\tau}^{des}=[\tau_{y}^{des}\,-\tau_{x}^{des}]$ is computed and integrated at each time step, accumulating into the desired CAM as follows,}
\begin{equation}
{h}_i^{des} = \int{\tau_{i}^{des}}\,dt,\quad i = x,y 
\label{eq/tauToCAM}
\end{equation}
{Here, the z-axis component is set to zero, and the desired CAM $\mathbf{h}^{des}\in\mathbb{R}^3$ is realized through the HQP-based WBIK described in Section \ref{Section/WBIK}.}

\section{Stepping Controller based on CP--MPC}
\label{Section/Stepping Controller}
While footstep position adjustment has been extensively studied in MPC approaches \cite{herdt2010online, joe2018balance, romualdi2022online, ding2022dynamic, kim2023foot}, optimizing the step time in the MPC formulation is still considered a challenging problem due to its non-convex nature. However, since step time adjustment improves balancing performance compared to adjusting footstep position alone, various methods such as heuristic approaches \cite{griffin2017walking, ding2022dynamic} or optimization based on LIPM assumptions \cite{kryczka2015online, nazemi2017reactive, jeong2019robust, khadiv2020walking, kim2023foot} have been conducted in addition to MPC approaches.

In this study, we developed a hierarchical control structure of CP--MPC and the stepping controller, enabling step time optimization based on MPC variables. Specifically, CP--MPC utilizes ZMP, CAM, and footstep adjustment to track the CP trajectory, while the stepping controller determined the footstep position and step time to control the CP offset at the end of the step.

\subsection{Overview of the Stepping Controller}
In this section, an overview is provided for the adjustment of footstep position and step time to control the CP offset in the stepping controller.
Our stepping controller is designed based on the QP optimization proposed by Khadiv et al. \cite{khadiv2016step, khadiv2020walking} using LIPFM-based CP end-of-step dynamics. The dynamics of CP end-of-step based on LIPFM can be expressed as follows, 
\begin{equation}
\label{eq/stepping_cpeos}
\bm{\xi}_{T}=\mathbf{f}^{} + \mathbf{b} = (\bm{\xi}-\mathbf{P}_{ssp})e^{-\omega t}\gamma^{}_{} + \mathbf{P}_{ssp}. 
\end{equation}
In \cite{khadiv2016step, khadiv2020walking}, the stepping controller is primarily designed to adjust the next footstep position $\mathbf{f}=[f_x\,f_y]$ and step time term $\gamma=e^{\omega T}$ ensuring that the CP offset $\mathbf{b}=[b_x\,b_y]$ closely aligns with the planned CP offset. The CP offset in LIPM-based walking has been identified as a critical factor that determines the initial point of the exponential growth of the CP in the next step, and has been emphasized in numerous studies \cite{englsberger2011bipedal, khadiv2016step, jeong2017biped, jeong2019robust, khadiv2020walking} as a key factor that affects the walking stability. Based on these concepts, we also adjust the footstep position and step time to control the CP offset.

Fig. \ref{figure/Steppingcontrol} presents the overall concept of our approach in the presence of disturbances. When the disturbance is applied to the robot, the CP end-of-step (purple point) $\bm{\xi}^{'}_{T}$ is predicted using (\ref{eq/stepping_cpeos}) to deviate significantly due to the disturbance. In this case, the stepping controller adjusts the footstep position $\mathbf{f}$ and the step time term $\gamma$ to control the CP offset $\mathbf{b}$ at the end of the step. 
\subsection{Parameter Decision based on CP--MPC}
\label{Section/Stepping Controller/Parameter decision}
\begin{figure}[t]
      \centering
      \includegraphics[width=0.95\linewidth]{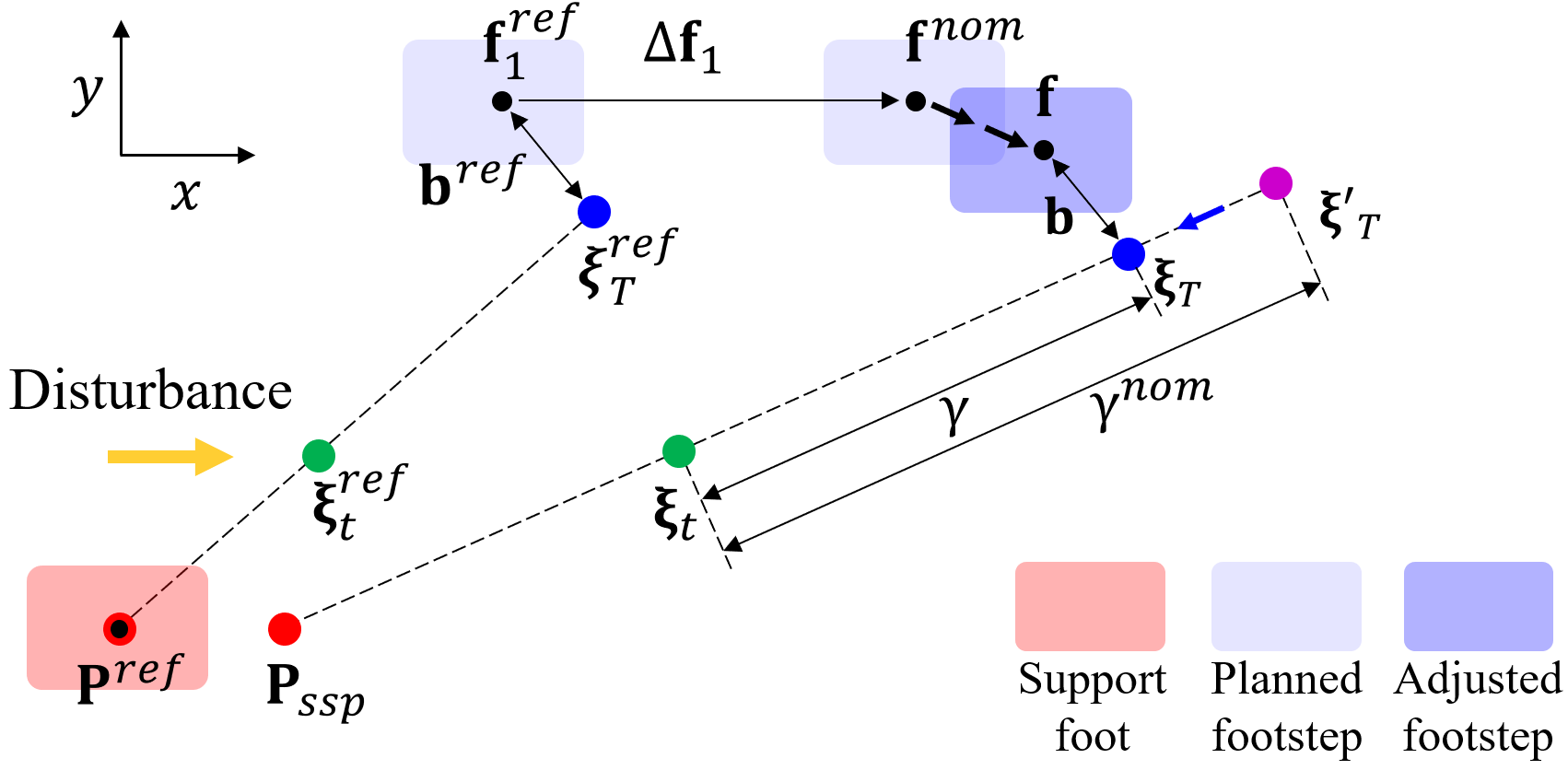} 
      \caption{Schematic representation of the stepping control; In order to control the CP offset, footstep position and step time are adjusted based on CP end-of-step dynamics.} 
      \label{figure/Steppingcontrol}
\end{figure}
This section describes the parameter decision for the nominal values of the next foot position $\mathbf{f}$, CP offset $\mathbf{b}$, step time term $\gamma$, and CMP parameters $\mathbf{P}_{ssp}$ in (\ref{eq/stepping_cpeos}).
The nominal footstep position $\mathbf{f}^{nom}=[{f}^{nom}_{x}\,{f}^{nom}_{y}]$ is defined as the sum of the reference next footstep position $\mathbf{f}_1^{ref}$ determined by the footstep planner and the footstep adjustment $\Delta \mathbf{f}_1$ optimized by CP--MPC to minimize the CP error over the future horizon, as follows, 
\begin{equation}
\mathbf{f}^{nom} = \mathbf{f}_1^{ref} + \Delta \mathbf{f}_1. 
\label{eq/F_nom}
\end{equation}
Here, $nom$ and $1$ refer to the nominal value and the next footstep, respectively.

Next, the nominal CP offset $\mathbf{b}^{nom}=[{b}^{nom}_x\,{b}^{nom}_y]$ is defined as the difference between the reference CP end-of-step $\bm{\xi}_{T}^{ref}=[{\xi}^{ref}_{T,x}\,{\xi}^{ref}_{T,y}]$ planned by the walking pattern generator and the reference next footstep position $\mathbf{f}^{ref}_1$, {which is equivalent to reference CP offset $\mathbf{b}^{ref}$}, to adhere to the target velocity command. 
\begin{equation}
\mathbf{b}^{nom} = \bm{\xi}^{ref}_{T} - \mathbf{f}^{ref}_1 = \mathbf{b}^{ref}. 
\label{eq/b_nom}
\end{equation} 
The nominal step time term $\gamma^{nom}=e^{\omega T^{ref}}$ is determined by a pre-defined step time based on the target velocity command.

{The CMP parameters $\mathbf{P}_{ssp}=[p_{ssp,x}\,p_{ssp,y}]$ (as shown in Fig. \ref{figure/Steppingcontrol}) are calculated as a single point accurately representing CP--MPC control inputs (ZMP and centroidal moment control inputs) corresponding to the remaining SSP within the prediction horizon. This approach enables a more precise reflection of the future behaviors of CP--MPC. In contrast to assuming that the CMP is fixed either at the center of the supporting foot or at the current position during the SSP, this method allows for a more accurate prediction of the CP end-of-step based on the inputs of CP--MPC.} The equation of the CMP parameter is as follows, 
\begin{equation}
{
\mathbf{P}_{ssp} = \frac{1}{1-A^{T_k-k}}\sum_{i=k}^{T_k-1}A^{T_k-i-1}\mathbf{B} 
\begin{bmatrix}
\mathbf{z}_{i} \\ \bm{\tau}_{i}
\end{bmatrix}}
\label{eq/stepping_CMP input}
\end{equation}
where $T_k$ represents the time step at which SSP ends within the MPC horizon, $\mathbf{z}_i=[z_{x,i}\,z_{y,i}]$ and $\bm{\tau}_i=[\tau_{y,i}\,-\tau_{x,i}]$ denote the ZMP and centroidal moment control input at each time step $i$, respectively. {The derivation for $\mathbf{P}_{ssp}$ is provided in Appendix \ref{section/appendix}}.
 
\subsection{Formulating a QP problem for Stepping Control}
\label{Section/Stepping Controller/QP formulation}
The QP formulation for adjusting the footstep position and the step time to control the CP offset is as follows.
\begin{align}
\label{eq/stepping_cost function}
& \underset{\mathbf{f}, \gamma, \mathbf{b}}{\text{min}} & & {\mathbf{w}_{f}}\| \mathbf{f}^{}-\mathbf{f}^{nom}\|^{2} + {w_{\gamma}}(\gamma-\gamma^{nom})^{2} \\ 
& & +&{\mathbf{w}_{b}}\|\mathbf{b}-\mathbf{b}^{nom}\|^{2} \nonumber \\ \label{eq/stepping_constraint}
& \text{s. t.} & & \underline{\mathbf{f}}^{}\leq  \mathbf{f}^{} \leq \overline{\mathbf{f}}^{}\\ 
& & &\underline{\gamma}^{}_{}\leq \gamma^{}_{} \leq \overline{\gamma}^{}_{} \nonumber\\
& & &\underline{\mathbf{b}}\leq \mathbf{b} \leq \overline{\mathbf{b}} \nonumber
\\ 
& & & \mathbf{f}^{} + \mathbf{b} = (\bm{\xi}-\mathbf{P}_{ssp})e^{-\omega t}\gamma^{}_{} + \mathbf{P}_{ssp} \nonumber 
\end{align}

The cost function (\ref{eq/stepping_cost function}) is composed of the footstep position error term, step time error term, and CP offset error term, and each cost is weighted by $\mathbf{w}_f$, $w_\gamma$, and $\mathbf{w}_b$, respectively. In each of the optimization variables, $\mathbf{f}=[f_x\,f_y]$ represents the next footstep position, $\gamma$ represents the step time term, and $\mathbf{b}=[b_x\,b_y]$ represents the CP offset. The variable $\gamma = e^{\omega T}$ is introduced to linearize the equality constraint in (\ref{eq/stepping_constraint}). The weighting parameter was set primarily to adjust the step time $T$ while also enhancing the consistency of $\mathbf{f}$ with the CP--MPC output. To achieve this, we set a smaller weighting value for the step time than the other weighting values as listed in Table. \ref{table/Parameters}.

In (\ref{eq/stepping_constraint}), the inequality constraint is composed of the upper and lower bounds for each optimization variable, and the equality constraint is defined by the LIPFM-based CP end-of-step dynamics in (\ref{eq/stepping_cpeos}).
The variable $\bm{\xi}$ represents the current CP, $t$ denotes the elapsed time since the start of the swing phase, and $\mathbf{P}_{ssp}$ refers to the CMP parameter used in the equality constraint.

\subsection{Characteristic of Our Parameter Selection Approach compared to Previous Methods}
\label{Section/Stepping Controller/Characteristic}
In this section, we introduce improvements in our parameter selection approach compared to previous studies \cite{khadiv2016step, nazemi2017reactive, khadiv2020walking, kim2023foot}.

Controlling the CP offset is crucial as it determines the initial velocity of the CP in the next step, thereby significantly affecting walking stability. Previous studies \cite{khadiv2016step, khadiv2020walking, nazemi2017reactive} rigorously controlled the CP offset. However, in these studies, the CP offset was derived under the assumption of a constant walking velocity, as follows,
\begin{align}
\label{eq/cp_offset_bolt_x}
{b}^{nom}_x &= \frac{L}{e^{\omega T^{nom}}-1},\\ {b}^{nom}_y &= (-1)^{n}\frac{D}{e^{\omega T^{nom}}-1}-\frac{W}{e^{\omega T^{nom}}-1}. 
\label{eq/cp_offset_bolt_y}
\end{align}
Here, $L$ represents the step length, $D$ denotes the default step width during walking, and $W$ represents the deviation with respect to the default step width. 
When $n$ is equal to 1, it indicates the right foot support phase, and when $n$ is equal to 2, it indicates the left foot support phase.

The nominal CP offset in (\ref{eq/cp_offset_bolt_x}) and (\ref{eq/cp_offset_bolt_y}) can be derived simply based on the CP end-of-step dynamics, and a detailed derivation is presented in \cite{khadiv2020walking}. However, these derivations include the assumption that the current step and the next step have the same values for $L$, $W$, and $T^{nom}$. Therefore, this assumption is violated when the walking velocity between the current and next steps are different (e.g., walking with different speeds for each step or sudden changes in walking direction). On the contrary, in our approach, while using the same QP formulation (\ref{eq/stepping_cost function}) as in previous studies, the CP offset is determined from (\ref{eq/b_nom}), taking into account the variations in walking velocity at each step. 


Next, in previous studies \cite{khadiv2016step, nazemi2017reactive, khadiv2020walking, kim2023foot}, it was assumed that the ZMP or CMP is fixed at the center of the support foot ($\mathbf{P}_{ssp} = \mathbf{P}^{ref}$), based on the point foot assumption. However, in most humanoid robots, the foot has a finite size and the ground reaction forces do not act exactly at the center of the support foot. 
Indeed, the larger the robot's feet, the further the robot's ZMP or CMP can deviate from the center of its support foot. This can potentially violate the point foot assumption on the CP dynamics based on LIPM or LIPFM. Therefore, in this paper, the CMP parameter obtained from CP--MPC is used to consider the behavior of CMP by CP--MPC instead of the point foot assumption.

In Section \ref{Section/Results/Simulation/SteppingController}, the effectiveness of the parameter selection approach is analyzed through a comparison with the previous methods.
{
\section{HQP--based Whole-Body Inverse Kinematics}
\label{Section/WBIK}
This section introduces an HQP--based WBIK to achieve the outputs from the walking pattern generator, CP--MPC, and stepping controller. The HQP-based WBIK extends our previous HQP--CAM controller \cite{kim2022humanoid}, which was limited to upper-body CAM control, into a whole-body control framework. 

The main objective of HQP--WBIK is to track the reference foot and CoM trajectories, as well as the desired CAM, while integrating initial pose recovery after CAM tracking motion into the hierarchical control framework. HQP--WBIK prioritizes CAM control over initial pose recovery by establishing strict task hierarchies to enhance balance performance.

The formulation of the HQP--WBIK is defined as follows.
\begin{align}
\label{eq/1st WBIK cost}
& \underset{\mathbf{\dot{q}_1}}{\text{min}}  & & \sum_j{ 
\| \mathbf{J}_{1,j}\mathbf{\dot{q}}_{1} -\mathbf{\dot{x}}^{des}_{1,j}\|^{2}_{\mathbf{w}_{\mathbf{J}}} +
\| \mathbf{A}\mathbf{\dot{q}}_{1}-\mathbf{h}^{des}\|^{2}_{\mathbf{w}_{\mathbf{A}}} +\| \mathbf{\dot{q}}_{1}\|^{2}_{\mathbf{w}_{\mathbf{\dot{q}}}}
} \\
\label{eq/1st WBIK constraint}
& \text{s. t.} & & \underline{\mathbf{\dot{q}}}_{1}\leq  \mathbf{\dot{q}}_{1} \leq \overline{\mathbf{\dot{q}}}_{1} \\ 
& & &  \mathbf{K}_q (\underline{\mathbf{q}}_{1} - \mathbf{q}_{1}) \leq  \dot{\mathbf{q}}_{1} \leq \mathbf{K}_q (\overline{\mathbf{q}}_{1} - \mathbf{q}_{1}) \nonumber \\
\label{eq/2nd WBIK cost}
& \underset{\mathbf{\dot{q}_2}}{\text{min}} & &\sum_j{\| \mathbf{J}_{2,j}\mathbf{\dot{q}}_{2} -\mathbf{\dot{x}}^{des}_{2,j}\|^{2}_{\mathbf{w}_{\mathbf{J}}}} 
+ \| \mathbf{\dot{q}}_{2}\|^{2}_{\mathbf{w}_{\mathbf{\dot{q}}}} \\
\label{eq/2nd WBIK constraint}
& \text{s. t.} & &  \mathbf{J}_{1,j} \mathbf{\dot{q}}_{2} = \mathbf{J}_{1,j} \mathbf{\dot{q}}_{1} \\
& & &  \mathbf{A} \mathbf{\dot{q}}_{1} - \mathbf{\boldsymbol{\varepsilon}} \leq \mathbf{A} \mathbf{\dot{q}}_{2} \leq \mathbf{A} \mathbf{\dot{q}}_{1} + \mathbf{\boldsymbol{\varepsilon}}   \nonumber \\
& & & \underline{\mathbf{\dot{q}}}_{2}\leq  \mathbf{\dot{q}}_{2} \leq \overline{\mathbf{\dot{q}}}_{2} \nonumber   \\ 
& & &  \mathbf{K}_q (\underline{\mathbf{q}}_{2} - \mathbf{q}_{2}) \leq  \dot{\mathbf{q}}_{2} \leq \mathbf{K}_q (\overline{\mathbf{q}}_{2} - \mathbf{q}_{2}) \nonumber 
\end{align} 
The optimization variable for the hierarchy~\(i\in\{1,2\}\) is the joint velocity \(\mathbf{\dot{q}}_i\in \mathbb{R}^{n+6}\), where \(n\) and \(6\) represent the number of actuated joints and virtual joints, respectively.
$\mathbf{J}_{i,j}$ and $\mathbf{\dot{x}}^{des}_{i,j}$ denote the Jacobian matrix and the desired velocity command for \(j\)-th task in the hierarchy $i$, respectively. 
The desired velocity command is defined as $\mathbf{\dot{x}}^{des}_{i,j} = \mathbf{\dot{x}}^{ref}_{i,j} + \mathbf{K}_p({\mathbf{x}}^{ref}_{i,j} - {\mathbf{x}}_{i,j})$ to minimize position or orientation error of the \(j\)-th task in the hierarchy $i$, where $\mathbf{K}_p$ is a positive feedback gains.
In the first hierarchy, the cost function~(\ref{eq/1st WBIK cost}) is composed of three cost terms.
The first cost term serves as the desired velocity command tracking control, ensuring that the robot follows the reference CoM position $\mathbf{c}^{ref}\in\mathbb{R}^3$ and desired foot pose as $\mathbf{e}^{des}_{L,R} = \mathbf{e}^{ref}_{L,R} + \triangle\mathbf{e}_{L,R}\in\mathbb{R}^6$.
The second cost term serves as the desired CAM tracking control.
Here, \(\mathbf{A}\in\mathbb{R}^{3 \times (n+6)}\) denotes the angular component of the centroidal momentum matrix, and \(\mathbf{h}^{des}\in\mathbb{R}^3\) is the desired CAM calculated from the CP--MPC.
The third cost term regulates the optimization variable~\(\mathbf{\dot{q}}_{1}\).
Each term is weighted by its respective positive diagonal weighting matrix:~\({\mathbf{w}_{\mathbf{J}}}\), \({\mathbf{w}_{\mathbf{A}}}\), and \({\mathbf{w}_{\mathbf{\dot{q}}}}\).
The constraints in~(\ref{eq/1st WBIK constraint}) consist of two inequality conditions, which limit the joint velocity~\(\mathbf{\dot{q}}_1\) and joint position~\(\mathbf{q}_1\) within their respective boundaries. 
The vectors $\overline{\mathbf{\dot{q}}}_{1}\in \mathbb{R}^{n+6}$ and $\underline{\mathbf{\dot{q}}}_{1}\in \mathbb{R}^{n+6}$ represent upper and lower bounds for joint velocities, respectively.
$\overline{\mathbf{{q}}}_{1}\in \mathbb{R}^{n+6}$ and $\underline{\mathbf{{q}}}_{1}\in \mathbb{R}^{n+6}$ represent upper and lower bounds for joint positions, where $\mathbf{K}_q$ is a positive diagonal matrix.

In the second hierarchy, the cost function in (\ref{eq/2nd WBIK cost}) consists of two cost terms, similar to those in~(\ref{eq/1st WBIK cost}).
The first cost term serves as the desired velocity command tracking control, required to maintain the initial pose such as pelvis orientation, chest orientation, and both hand poses. 
The constraints in (\ref{eq/2nd WBIK constraint}) include three inequality conditions and one equality condition.
The first and second constraints are set to ensure a strict task hierarchy, where lower-priority tasks do not affect higher-priority tasks.
If $\mathbf{\boldsymbol{\varepsilon}} = 0$, CAM control has higher priority than the initial pose recovery since the optimal solution of (\ref{eq/2nd WBIK cost}) does not affect the result from the first QP (\ref{eq/1st WBIK cost}).
However, it is impossible to generate the return motion without violating the prior CAM control calculated from the higher-level hierarchy.
To address this problem, a tolerance $\mathbf{\boldsymbol{\varepsilon}}$ was introduced in \cite{kim2022humanoid} to relax the strict hierarchy by a specified magnitude.
The tolerance $\mathbf{\boldsymbol{\varepsilon}}$ is adjusted online according to the 2-norm of the desired CAM.
{Lastly, the optimized $\mathbf{\dot{q}}$, excluding the underactuated virtual joint, becomes $\mathbf{\dot{q}}^{des}$, while $\mathbf{{q}}^{des}$ is obtained by integrating $\mathbf{\dot{q}}^{des}$ from the current time step.} 

{Note that in our framework, we assigned a higher weighting parameter to maintain the initial pelvis pose compared to the hand and chest pose maintenance, which is a design choice to limit excessive pelvic movement under small disturbances.}
}
\begin{figure*}[t]
      \centering  
      \includegraphics[width=0.99\linewidth]{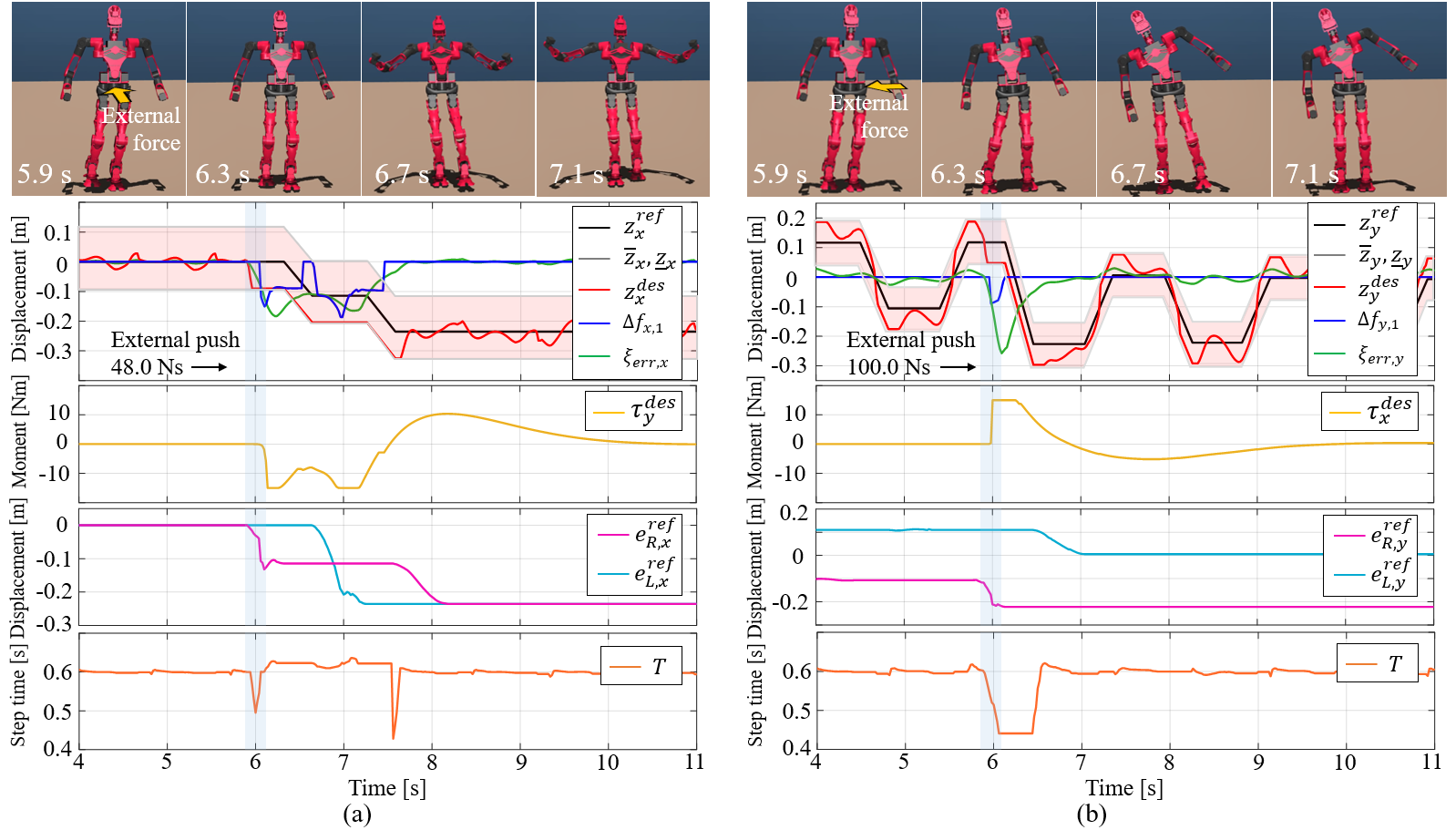} 
      \caption{{Simulation results are presented for the robot's response to external forces applied along both the negative x- and y-directions, shown in (a) and (b) respectively.}} 
      \label{figure/Simulation_1_Snapshot_Graph_}
\end{figure*}
\section{Results of Simulations and Experiments}
\label{Section/Results} 
\subsection{System Overview}
\label{Section/Results/System overview}
This section provides a system overview for both simulations and real robot experiments. The humanoid robot, TOCABI, utilized in the simulations and experiments comprises a total of 33 degrees of freedom (DOFs): 16 for the arms, 12 for the legs, 3 for the waist, and 2 for the neck. Its physical dimensions are approximately 1.8 m in height, weighing around 100 kg, and with a foot size of 15 cm $\times$ 30 cm. The upper-body actuators comprise Parker BLDC motors with harmonic gears, while the lower-body actuators incorporate Kollmorgen BLDC motors with harmonic gears.
The current control of the robot is performed by Elmo Motion Control's Gold Solo Whistle servo controller, and the communication between the servo controller and the main PC is achieved via EtherCAT communication.
MicroStrain's 3DM-GX5-25 IMU is attached to the pelvis, and ATI's mini85 F/T sensor is mounted on each foot. The algorithmic operation and torque command frequency of the robot are set to 2 kHz, and the walking pattern generator and CP--MPC operate at 50 Hz, due to the high computational load involved, through parallel threads. The walking pattern generator and CP--MPC have respective MPC time horizons of 2.5 s and 1.5 s. It should be noted that the positive direction of the x-axis represents the robot's forward direction, while the positive direction of the z-axis is the opposite direction of the gravity vector. The simulator employed in this study is the MuJoCo simulator \cite{todorov2012mujoco}. {The qpOASES \cite{ferreau2014qpoases} was employed to solve QP optimization problems, while RBDL \cite{felis2017rbdl} was utilized for calculating the kinematics and dynamics of the robot.} 

\begin{figure*}[t]
      \centering 
      \includegraphics[width=0.98\linewidth]{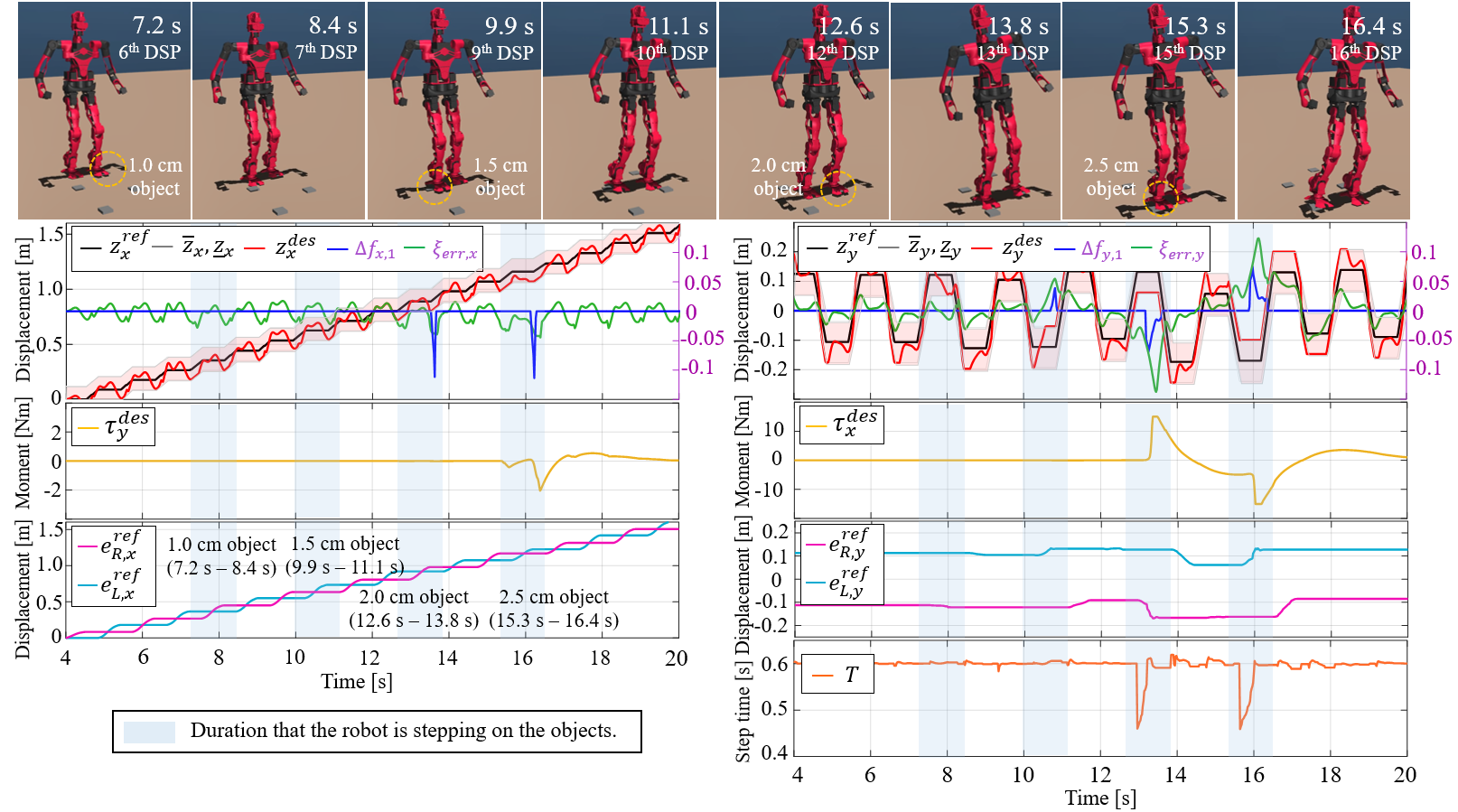} 
      \caption{{Simulation results are presented for the robot's forward walking while overcoming four unexpected objects.}}
      \label{figure/Simulation_1_2_uneven_Snapshot_Graph_}
\end{figure*}

\subsection{{Simulations to Validate Robustness of the Proposed Method under External Forces}}
\label{Section/Results/Simulation/extforce}

Simulations were conducted to evaluate the balancing capability of the robot against external forces using the proposed method. The simulations involved a scenario where the robot walked in the place and external forces are applied to the robot's pelvis link in the x- and y-directions during an SSP (left foot support). The step duration was configured to consist of an SSP of 0.6 s and a DSP of 0.3 s. 

In the first simulation, an external force of 240 N was applied to the robot's pelvis in the negative x-direction for 0.2 s. Fig. \ref{figure/Simulation_1_Snapshot_Graph_}(a) shows snapshots and data of the robot's response to the external force. 
{The snapshots illustrate the overall balancing process, where the robot maintains balance through a combination of ZMP control (ankle strategy), CAM control (hip strategy), and back steps (stepping strategy).}

{The graph in Fig. \ref{figure/Simulation_1_Snapshot_Graph_}(a) represents the outputs of the CP--MPC and stepping controller in the x-direction. 
At approximately 5.9 s, the robot was pushed backward, causing the CP error $\xi_{err,x}$ (green line) to increase in the negative x-direction due to the external force. To reduce the CP error, the desired ZMP $z_x^{des}$ (red line) is generated by the CP--MPC. Despite the attempts to reduce the CP error through the desired ZMP, it remained high due to the significant external force and the ZMP constraint of -9 cm. As a result, the CP error increased and reached up to -18.3 cm. Then, footstep adjustment $\Delta f_{x,1}$ (blue line) was generated to relax the ZMP constraint and the desired centroidal moment $\tau_y^{des}$ (yellow line) was also generated to further reduce the CP error. 
The desired ZMP was generated up to the relaxed ZMP constraint until approximately 7.6 s, while the centroidal moment was generated up to the limit of -15 Nm until approximately 7.2 s.
Meanwhile, the stepping controller performed a total of two back-steps (magenta and cyan line) based on $\Delta f_{x,1}$, moving approximately -23.5 cm in conjunction with an optimal step time $T$ (orange line). As a result, the CP error is reduced to approximately zero, and the robot maintains a balance against the external force.}

In the following simulation, an external force of 500 N was applied to the robot's pelvis in the negative y-direction for 0.2 s. Fig. \ref{figure/Simulation_1_Snapshot_Graph_}(b) presents snapshots and y-direction data of the robot's response to the external force. Similar to the previous simulation, the robot maintained balance by stepping in the direction of the external force while performing ZMP control and CAM control. 

Fig. \ref{figure/Simulation_1_Snapshot_Graph_}(b) represents the robot's data, including the CP--MPC and stepping controllers' y-direction outputs. 
{At approximately 5.9 s, the external force was applied to the robot in the negative y-direction, leading to a significant deviation in the CP error $\xi_{err,y}$. The CP--MPC attempted to decrease the increasing CP error by generating the desired ZMP $z_y^{des}$ up to the ZMP constraint. At approximately 6 s, footstep adjustment of -8.6 cm $\Delta f_{y,1}$ was generated to relax the ZMP constraint, while a desired centroidal moment $\tau_x^{des}$ of 15 Nm was also generated to reduce the CP error. Based on the generated $\Delta f_{y,1}$, the stepping controller adjusted the swing foot trajectory and reduced the step time $T$ to allow for stepping at approximately 6.1 s. Subsequently, continuous CP control using each strategy reduced the CP error from approximately -25.7 cm initially, gradually decreasing until the robot was able to maintain balance against external forces.}

\subsection{{Simulation to Validate Robustness of the Proposed Method on Uneven Terrain}}
\label{Section/Results/Simulation/uneven}
In this simulation, the robot walks over objects placed on the ground to validate the balancing performance of the proposed method on uneven terrain. Four objects, each with a length and width of 5 cm, were positioned in front of the robot with a distance of 30 cm between them. To make the robot step on each object alternately, two objects were placed on the left and two on the right with respect to the robot's center. 
The thickness of the objects ranged from 1.0 cm to 2.5 cm with an increase of 0.5 cm. The robot walked a distance of 2.0 m with a step length of 0.1 m, and the step duration consisted of 0.6 s of SSP and 0.3 s of DSP. 

Fig. \ref{figure/Simulation_1_2_uneven_Snapshot_Graph_} shows simulation snapshots and the corresponding output data. 
In the snapshots, the robot was able to navigate forward while stepping on the objects in its path. The first and second objects were overcome using ZMP and stepping control (ankle and stepping strategies). For the third and fourth objects, CAM control (hip strategy) was additionally employed to navigate past them. The analysis of this process is explained based on the data presented in the graphs. 
Note that in the second row of Fig. \ref{figure/Simulation_1_2_uneven_Snapshot_Graph_}, the purple right y-axis highlights the CP error $\xi_{err}$ and footstep adjustment $\Delta \mathbf{f}_{1}$ to enhance the visibility of their variations.

{First, at approximately 7.2 s, which is the start of the 6th DSP, the robot steps on the object with a thickness of 1.0 cm diagonally in the y-direction, causing external disturbances.
However, the robot was not significantly disturbed by the object of relatively low thickness, and the reduction in CP error was solely achieved by generating the desired ZMP until approximately 8.4 s when the robot passed through the object.}

{At the start of the 9th DSP, approximately 9.9 s, the robot encountered the 1.5 cm object. To overcome the resulting disturbance, the CP--MPC generated a footstep adjustment $\Delta f_{y,1}$ of 3.8 cm at approximately 10.8 s, in contrast to the initial encounter with 1.0 cm object. By utilizing $\Delta f_{y,1}$ as a nominal value in the stepping controller, the reference swing foot position $e^{ref}_{L,R,y}$ was adjusted, enabling the robot to overcome the disturbance.}

{Next, at the start of the 12th DSP, approximately 12.6 s, the robot stepped on the 2 cm object. To alleviate the CP errors, the desired ZMP was generated up to the constraint in both the x- and y-directions. Additionally, $\Delta f_{y,1}$ of -7.0 cm was generated at approximately 13.2 s and $\Delta f_{x,1}$ of -11.2 cm was also generated at 13.5 s. Notably, the large CP error in the y-direction led to a centroidal moment $\tau^{des}_x$ of 15 Nm. The robot was able to overcome disturbances by adjusting footstep position and step time to a greater extent than in the second disturbance response, while also implementing CAM control.} 

{Finally, at the start of the 15th DSP, approximately 15.3 s, the robot stepped on the thickest object of 2.5 cm. 
Due to the large disturbance, the desired ZMP in the x- and y-directions was generated up to each ZMP constraint in three steps.
$\Delta f_{y,1}$ of 7.5 cm was generated at approximately 15.9 s, and $\Delta f_{x,1}$ was generated up to -11.4 cm at approximately 16.2 s. The centroidal moment $\tau^{des}_x$ in the x-direction was also generated up to the constraint. With footstep position adjustment, step time was also reduced by up to 0.46 s, allowing the robot to overcome the disturbance. In the x-direction, due to the larger range of ZMP control and footstep adjustments compared to the y-direction, only a small amount of centroidal moment $\tau^{des}_y$ was generated in overcoming all objects.}

In conclusion, this simulation showed that appropriate strategies were executed to mitigate disturbances of various magnitudes caused by uneven terrain.

\subsection{{Simulation to Evaluate Balancing Performance based on the Different Combinations of Each Balance Strategy}}
\begin{figure}[t]
      \centering
      \includegraphics[width=1.0\linewidth]{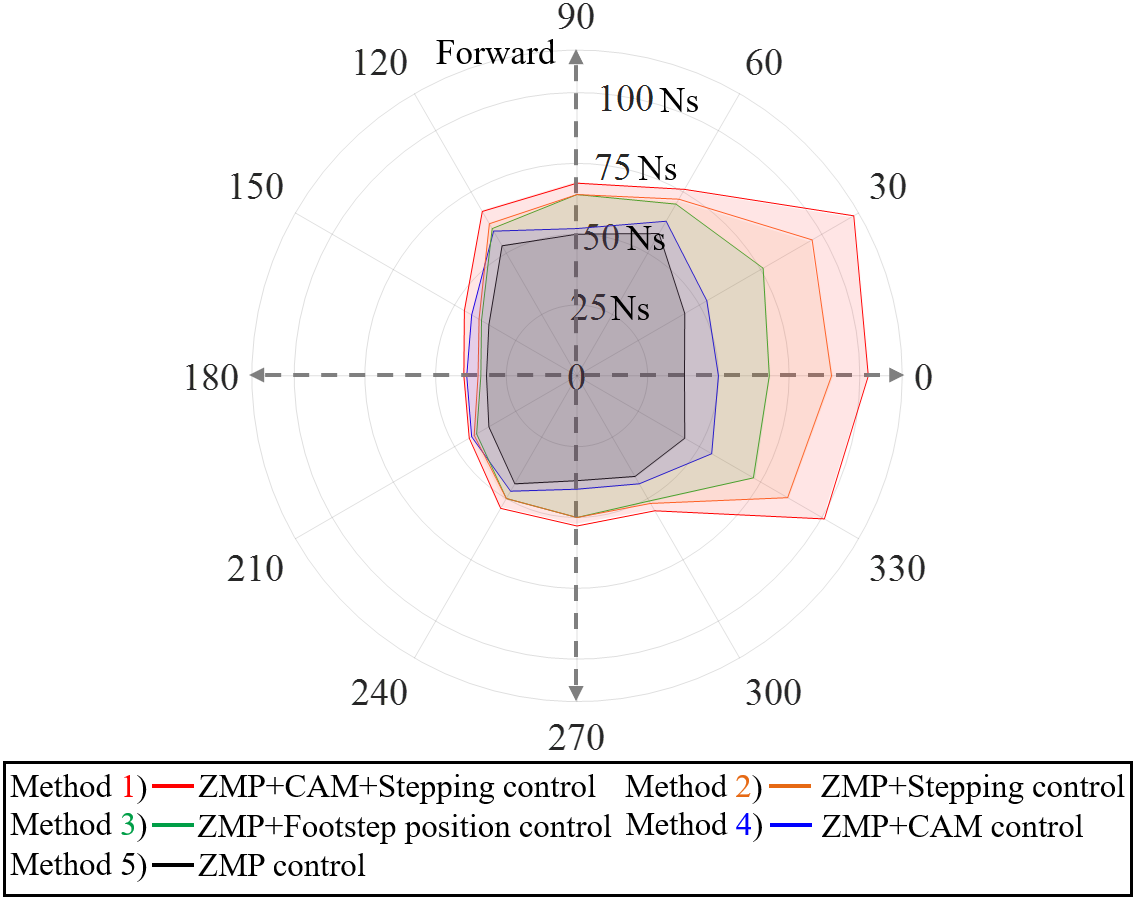} 
      \caption{{Comparisons of the maximum endurable impulse across different strategy combinations.}}
      \label{figure/Simulation1_DP}
\end{figure}

Comparative simulations were conducted to evaluate how the balancing performance varies based on different combinations of balancing strategies in response to disturbances. Presented below is a summary of the implementation methods employed to combine each strategy {(see Fig. \ref{figure/Simulation1_DP})}.
\begin{itemize}
    \item Method 2) is composed of a combination of ZMP control and stepping control (footstep position and step time adjustment), excluding the CAM control. To achieve this, 
    all terms related to the centroidal moment $\bm{\uptau}$ in (\ref{eq/CP-MPC cost function}) and (\ref{eq/CP-MPC constraint}) were excluded. 
    \item Method 3) has the same structure as method 2), but it does not consider adjusting the step time $T$ in the stepping controller. Therefore, only the next footstep position $\mathbf{f}$ is adjusted in the stepping controller.
    \item Method 4) is a combination of ZMP control and CAM control, excluding the stepping control. To achieve this, all terms related to $\Delta \mathbf{F}$ in (\ref{eq/CP-MPC cost function}) and (\ref{eq/CP-MPC constraint}) were excluded, and the stepping controller was also excluded.
    \item Method 5) only performs ZMP control. All terms related to $\uptau$ and $\Delta \mathbf{F}$ in (\ref{eq/CP-MPC cost function}) and (\ref{eq/CP-MPC constraint}) were excluded, as well as the stepping controller.
\end{itemize}
In order to analyze the robustness of each method against disturbances from various directions, the maximum external impulse that the robot could withstand was analyzed by varying the direction of the external force in increments of 30 deg.
Similar to previous simulations, an external force was applied to the pelvis of the robot for 0.2 s while the robot walked in place. In Fig. \ref{figure/Simulation1_DP}, the maximum external impulse that the robot can withstand is represented by the form of a polygon, which is referred to as the disturbance polygon (DP) in this paper. The direction of the external force vector starts from the robot and points outward, and a force of 90 deg represents a force that is directed from back to front (+x-axis).

As expected, method 5), which only tracks the desired ZMP, exhibited the lowest balancing performance compared to other methods. This is because CP control is achieved only through ZMP control, and ZMP control performance is limited within the support polygon.
Method 4), which tracks the desired ZMP and centroidal moment, extended the range of the control input that controls CP beyond the support polygon compared to method 5), resulting in a {15.8 $\%$} increase in the average maximum external impulses that the robot can withstand.
Method 3), which tracks the desired ZMP and adjusts the footstep position, significantly increases the maximum endurable impulses compared to methods 4) and 5) at 0, 30, and 330 degrees, where the support area is greatly expanded through footstep position adjustments.
However, when footstep position adjustment is restricted due to self-collision (i.e., near 180 degrees), its contribution to robustness decreases relative to other directions, becoming comparable to that of method 4).
In method 2), which simultaneously tracks the desired ZMP and adjusts footstep position and step time, it was possible to adjust the footstep position more quickly in response to disturbances compared to method 3). In particular, this led to a noticeable improvement in balancing performance, especially at 0, 30, and 330 degrees. 
Our proposed method, which combines all strategies, exhibited superior balancing performance across all directions compared to other sub-combinations. This is because all limbs of the robot (supporting leg, upper body, and swing leg) participated in CP control through the three balance strategies.
As a result, our method led to an increase of {14.5 $\%$, 27.7 $\%$, 48.8 $\%$, and 72.2 $\%$} in the average maximum external impulses that the robot could withstand compared to methods 2), 3), 4), and 5), respectively.
 
\subsection{{Simulation to Validate the Effectiveness of the Variable Weighting Method}}
\label{Section/Results/Simulation/VariableWeighting}
\begin{figure}[t]
      \centering
      \includegraphics[width=0.980\linewidth]{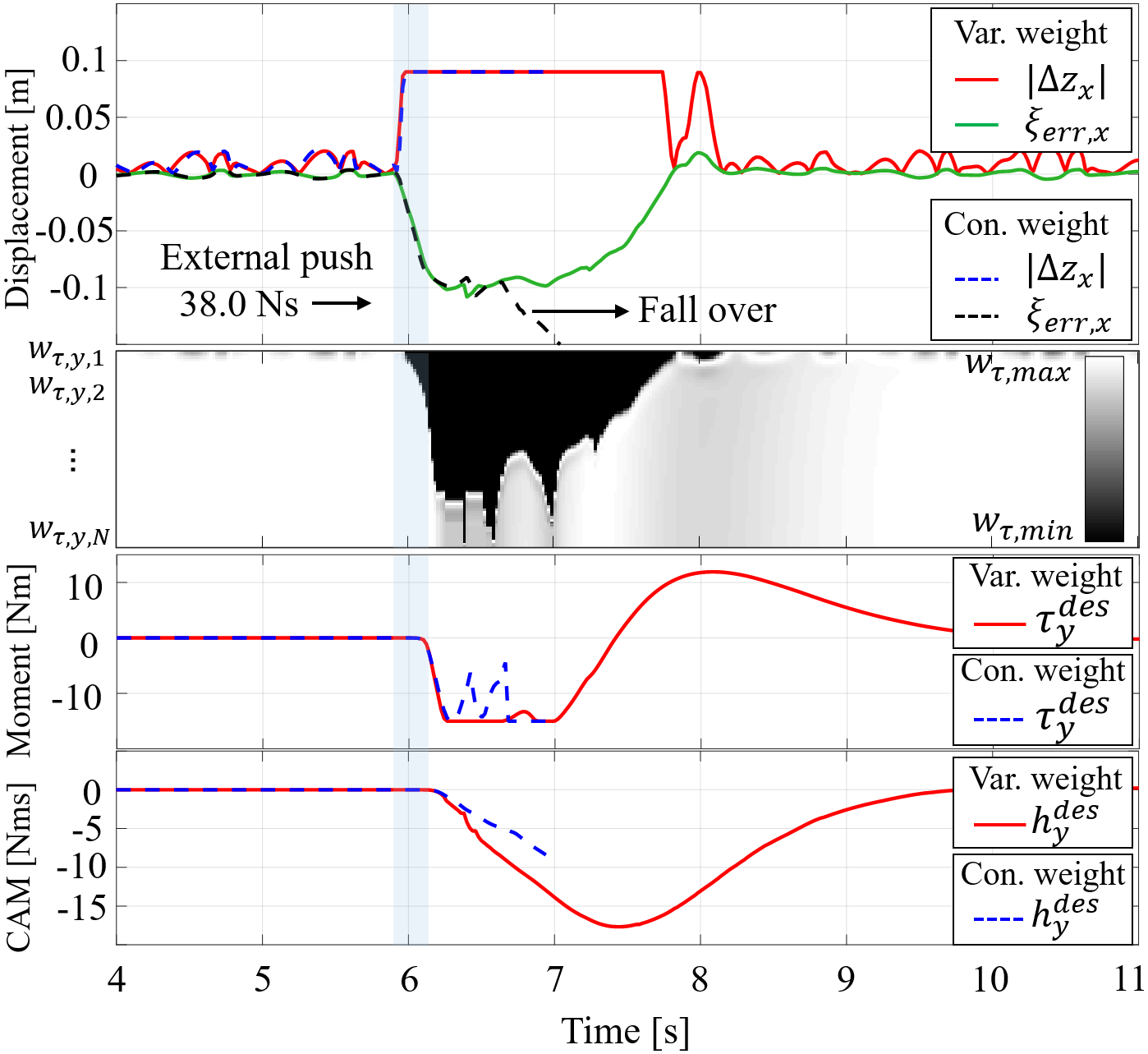} 
      \caption{{Impact of variable weight parameters on CAM generation and resulting balancing performance.}}
      \label{figure/Simulation_2_graph}
\end{figure} 
\begin{figure}[t]
      \centering
      \includegraphics[width=1.0\linewidth]{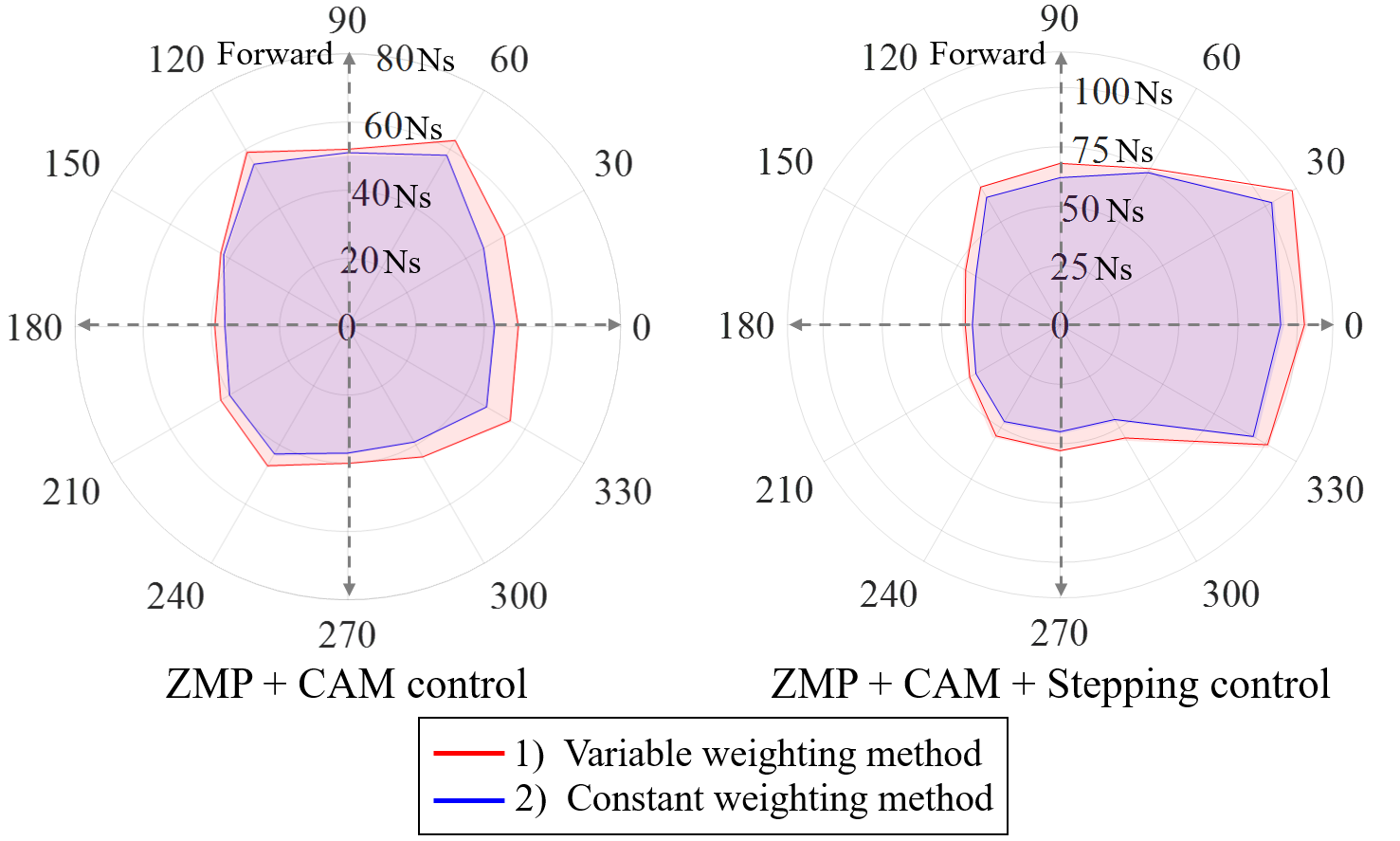} 
      \caption{{Comparisons of the maximum endurable impulse between variable and constant weighting method.}}
      \label{figure/Simulation_2_DP}
\end{figure} 
In order to evaluate the effectiveness of the variable weighting method in the presence of disturbance, a comparative simulation with the constant weighting method was performed. The simulation setup was identical to that in Sec \ref{Section/Results/Simulation/extforce}, where an external force was applied to the robot's pelvis in the negative x-direction during walking in place. 
{In this simulation, ZMP control and CAM control without stepping control were employed, and only the effects of external forces in the x-direction were analyzed. In the constant weighting method, $\mathbf{w}_{\tau}$ is set equal to $\mathbf{w}_{\tau,max}$ in the variable weighting method, aiming to restrict CAM generation for small disturbances, as in the variable weighting method.} 

Fig. \ref{figure/Simulation_2_graph} presents the results of comparing the performance of the two methods. The first graph shows the relationship between CP error and the magnitude of the current delta ZMP $|\Delta z_x|$. The delta ZMP refers to the additional ZMP input needed to control the CP error, which is calculated as the difference between the desired ZMP and the reference ZMP, as explained in Section \ref{Section/Variable Weighting}. When the external force is applied to the robot at approximately 5.9 s, both methods exhibit an increasing trend in CP error (green line and black dotted line), leading to an increase in the magnitude of delta ZMP input (red line and blue dotted line).  
Note that the weighting parameters changes according to $N$ series of delta ZMP inputs, but the graph depicts only the first element of delta ZMP inputs to show the overall behavior with respect to CP error. 

The second graph depicts the variation of $N$ weighting parameters (vertical axis) based on the magnitude of delta ZMP inputs $|\Delta \mathbf{Z}_x|$ using a color-map. In the variable weighting method, the graph appears white when no disturbance is applied to the robot. 
However, as the delta ZMP increases due to disturbances after approximately 6 s, the weighting parameters of the CAM damping cost decrease, causing the color-map to turn dark.
In contrast, the constant weighting method always has a constant weighting value regardless of delta ZMP inputs.

As evident from the third and fourth graphs, the difference between the two methods impacts the generation $\tau_y^{des}$ for disturbances. As a result, it directly affects the desired CAM $h^{des}_y$, which is generated by integrating $\tau_y^{des}$.
Although the delta ZMP was almost the same (as shown in the graph of the first row), using constant weighting method resulted in the insufficient generation of CAM compared to the variable weighting method. This led to the robot falling over after approximately 6.8 s when the constant weighting is used.

{The performance of the variable weighting method was compared with that of the constant weighting method using DP, and the results were illustrated in Fig. \ref{figure/Simulation_2_DP}. }
{With the variable weighting method, the average maximum external impulses that the robot can withstand increased by 9.88 $\%$ and 9.71 $\%$ in the ZMP+CAM method and the ZMP+CAM+Stepping method, respectively, compared to the constant weighting method.}

\subsection{Simulation to evaluate walking performance based on different parameter selections of the stepping controller}
\label{Section/Results/Simulation/SteppingController}
\begin{figure}[t]
      \centering
      \includegraphics[width=1.0\linewidth]{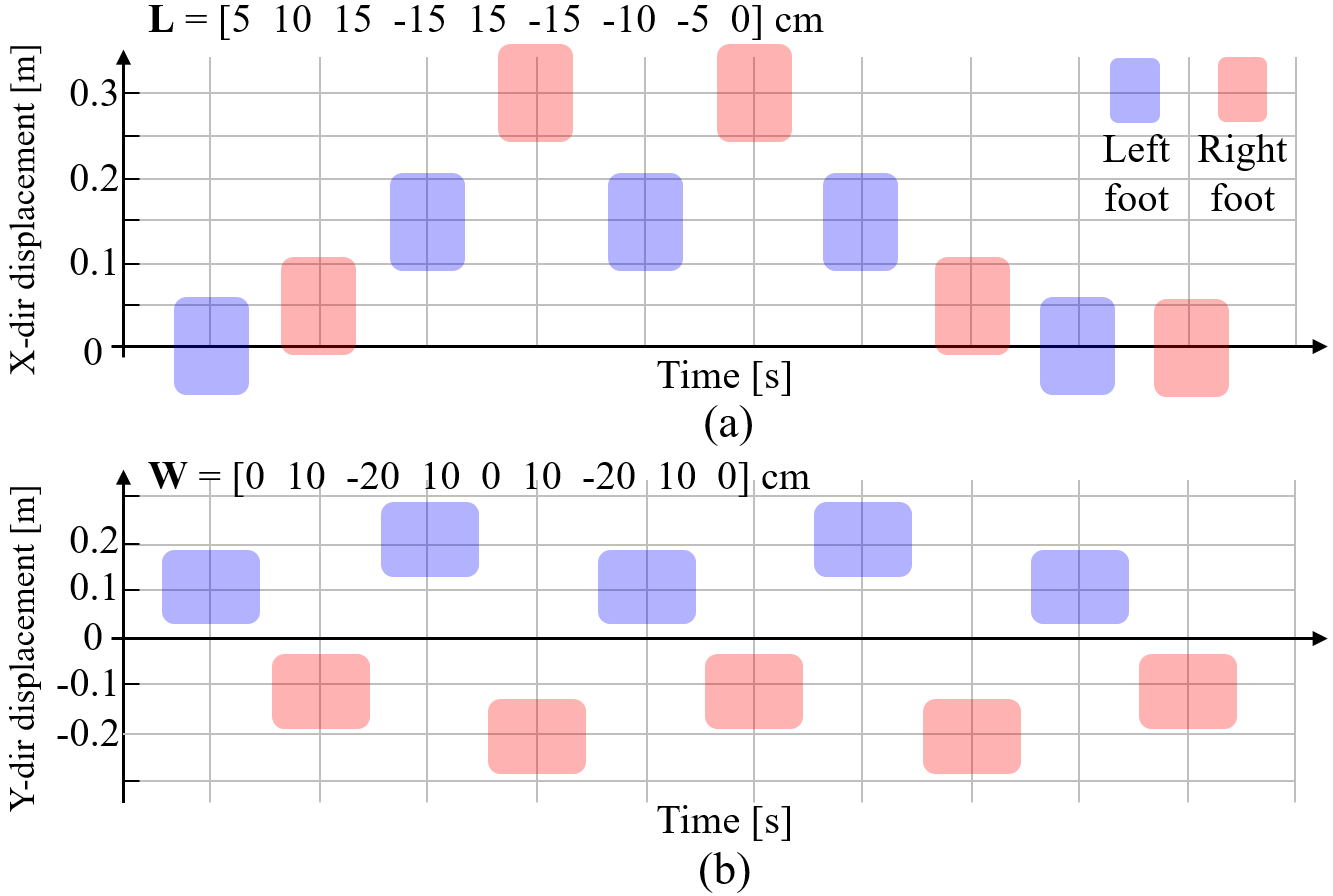} 
      \caption{Planned footstep placement used in the simulation of Section \ref{Section/Results/Simulation/SteppingController} (a) forward and backward walking at various speeds in the x-direction. (b) lateral walking while the robot doesn't move in the x-direction.}
      \label{figure/Simulation3_footstep}
\end{figure}

In this section, simulations were conducted to analyze the impact of parameter selection in the stepping controller on walking stability. Specifically, the CP tracking error $\bm{\xi}_{err}$ and CP offset error $\mathbf{b}_{err}$ were analyzed according to the two parameters (nominal CP offset $\mathbf{b}^{nom}$ and CMP parameter $\mathbf{P}_{ssp}$). Fig. \ref{figure/Simulation3_footstep} illustrates the walking scenarios conducted in this simulation. Fig. \ref{figure/Simulation3_footstep}(a) represents forward and backward walking with step length $L$, and in this case, the deviation of the step width $W$ is zero during walking. Fig. \ref{figure/Simulation3_footstep}(b) represents the lateral walking with step width deviation $W$, where the default step width $D$ is 20.5 cm. Additionally, in this case, the step length $L$ is zero during walking.

First, in order to analyze the effects of the nominal CP offset $\mathbf{b}^{nom}$ on walking stability, simulations were conducted in two walking scenarios depicted in Fig. \ref{figure/Simulation3_footstep}.
For comparative simulations, an analytical CP offset calculation method proposed in \cite{khadiv2016step, khadiv2020walking} was adopted. 
Here, both methods used the CMP parameter $\mathbf{P}_{ssp}$ calculated by (\ref{eq/stepping_CMP input}).
\begin{table}[]
\caption{{Analysis of CP offset errors and CP tracking errors according to the selection of nominal CP offset $\mathbf{b}^{nom}$.}}
\label{table/stepping b nom}
\centering
\resizebox{\columnwidth}{!}{%
\begin{tabular}{|l|c|cccc|}
\hline
\multicolumn{1}{|c|}{} &  & \multicolumn{4}{c|}{RMS error [cm]} \\ \cline{3-6} 
\multicolumn{1}{|c|}{\multirow{-2}{*}{Scenario}} & \multirow{-2}{*}{Method} & \multicolumn{1}{c|}{$\xi_x$} & \multicolumn{1}{c|}{$\xi_y$} & \multicolumn{1}{c|}{$b_x$} & $b_y$\\ \hline
 & \multicolumn{1}{l|}{\begin{tabular}[c]{@{}c@{}} {Analytical CP offset} \\ {calculation}  \end{tabular}} & \multicolumn{1}{c|}{1.27} & \multicolumn{1}{c|}{3.24} & \multicolumn{1}{c|}{4.04} & 6.47 \\ \cline{2-6} 
\multirow{-2}{*}{\begin{tabular}[c]{@{}l@{}}(a) Forward and \\ backward walking\end{tabular}} & Our approach & \multicolumn{1}{c|}{\cellcolor[HTML]{D7D7D7}0.81} & \multicolumn{1}{c|}{\cellcolor[HTML]{D7D7D7}0.98} & \multicolumn{1}{c|}{\cellcolor[HTML]{D7D7D7}1.14} & \cellcolor[HTML]{D7D7D7}0.66 \\ \hline
 & \multicolumn{1}{l|}{\begin{tabular}[c]{@{}c@{}} {Analytical CP offset} \\ {calculation} \end{tabular}} & \multicolumn{1}{c|}{0.47} & \multicolumn{1}{c|}{4.06} & \multicolumn{1}{c|}{1.37} & 8.23 \\ \cline{2-6} 
\multirow{-2}{*}{(b) Lateral walking} & Our approach & \multicolumn{1}{c|}{\cellcolor[HTML]{D7D7D7}0.24} & \multicolumn{1}{c|}{\cellcolor[HTML]{D7D7D7}1.40} & \multicolumn{1}{c|}{\cellcolor[HTML]{D7D7D7}0.14} & \cellcolor[HTML]{D7D7D7}1.05 \\ \hline
\end{tabular}%
}
\end{table}
Table \ref{table/stepping b nom} shows the root mean square (RMS) error of the CP offset and the CP tracking when different nominal CP offsets are selected in each scenario. When adopting $\mathbf{b}^{nom}$ calculated using (\ref{eq/cp_offset_bolt_x}) and (\ref{eq/cp_offset_bolt_y}), substantial CP tracking errors and CP offset errors were observed in both scenarios (a) and (b) compared to our method. As mentioned in Section \ref{Section/Stepping Controller/Characteristic}, this method assumes that both $L$ in (\ref{eq/cp_offset_bolt_x}) and $W$ in (\ref{eq/cp_offset_bolt_y}) remain constant for the current and the next step. However, as shown in Fig. \ref{figure/Simulation3_footstep}, this assumption is violated in scenario (a) where $L$ changes at each step, or in scenario (b) where $W$ changes at each step. As a result, unintended CP velocity is induced at the beginning of each step, leading to a deterioration in walking stability. 
When our method is used, the CP offset is calculated based on (\ref{eq/b_nom}) according to the walking velocity command without the constant walking velocity assumption. Consequently, the CP offset is controlled to follow the walking velocity command, allowing the robot to minimize CP tracking error and start the next step with a small CP error.

\begin{table}[t]
\caption{{Analysis of CP offset errors and CP tracking errors according to the selection of CMP parameter $\mathbf{P}_{ssp}$.}}
\label{table/stepping cmp}
\centering
\resizebox{\columnwidth}{!}{%
\begin{tabular}{|l|c|cccc|}
\hline
\multicolumn{1}{|c|}{} &  & \multicolumn{4}{c|}{RMS error [cm]} \\ \cline{3-6} 
\multicolumn{1}{|c|}{\multirow{-2}{*}{Scenario}} & \multirow{-2}{*}{Method} & \multicolumn{1}{c|}{$\xi_x$} & \multicolumn{1}{c|}{$\xi_y$} & \multicolumn{1}{c|}{$b_x$} & $b_y$\\ \hline
 & \multicolumn{1}{l|}{\begin{tabular}[c]{@{}c@{}}{Fixed CMP with} \\ point foot assumption\end{tabular}} & \multicolumn{1}{c|}{1.02} & \multicolumn{1}{c|}{1.35} & \multicolumn{1}{c|}{1.66} & 1.22 \\ \cline{2-6} 
\multirow{-2}{*}{\begin{tabular}[c]{@{}l@{}}(a) Forward and \\ backward walking\end{tabular}} & Our approach & \multicolumn{1}{c|}{\cellcolor[HTML]{D7D7D7}0.81} & \multicolumn{1}{c|}{\cellcolor[HTML]{D7D7D7}0.98} & \multicolumn{1}{c|}{\cellcolor[HTML]{D7D7D7}1.14} & \cellcolor[HTML]{D7D7D7}0.66 \\ \hline
 & \multicolumn{1}{l|}{\begin{tabular}[c]{@{}c@{}}{Fixed CMP with} \\ point foot assumption \end{tabular}} & \multicolumn{1}{c|}{0.43} & \multicolumn{1}{c|}{1.74} & \multicolumn{1}{c|}{0.38} & 1.60 \\ \cline{2-6} 
\multirow{-2}{*}{(b) Lateral walking} & Our approach & \multicolumn{1}{c|}{\cellcolor[HTML]{D7D7D7}0.24} & \multicolumn{1}{c|}{\cellcolor[HTML]{D7D7D7}1.40} & \multicolumn{1}{c|}{\cellcolor[HTML]{D7D7D7}0.14} & \cellcolor[HTML]{D7D7D7}1.05 \\ \hline
\end{tabular}%
}
\end{table}

Next, simulations were performed to analyze the impact of the CMP parameter $\mathbf{P}_{ssp}$ on walking stability for both scenarios (a) and (b). Here, our CP offset calculation method (\ref{eq/b_nom}) was used.
For the comparison simulations, the fixed CMP that assumed the point foot \cite{khadiv2016step, khadiv2020walking, kim2023foot} was compared with our method, and the results are presented in Table \ref{table/stepping cmp}. In both scenarios (a) and (b), the fixed CMP under the point foot assumption resulted in larger errors compared to our method. 
{The reason is that the point foot assumption assumes the CMP to act at the center of the foot during SSP, which fails to accurately reflect the CMP position for the robot with large feet. As a result, the CP end-of-step is inaccurately predicted, and CP offset control based on this prediction leads to a deterioration of walking stability.
On the other hand, our approach calculates the CMP parameter by accurately approximating series of CP--MPC control inputs during the SSP to a single point. This enables the accurate prediction of CP end-of-step by reflecting the tendencies of the CMP acting on the robot.}

In conclusion, it is important to adopt appropriate parameters for the stepping controller to enhance the overall stability of walking.


\subsection{{Simulations to Compare Robustness against Disturbances with QP-based CP controller [39]}}
\label{Section/Results/Simulation/Jeong}

In this section, simulations were conducted to compare the robustness against disturbances between the proposed method and state-of-the-art QP-based CP controller \cite{jeong2019robust}. For the comparative simulations, the algorithm proposed by \cite{jeong2019robust} was implemented in our walking control framework (see Fig. \ref{figure/overallframe}). In the implemented algorithm, the QP-based CP controller replaced our CP--MPC and stepping controller, while all other local controllers and trajectory planners described in Section \ref{Section/Overall Framework} remained unchanged.

The reasons for selecting the QP-based {CP controller} \cite{jeong2019robust} as a suitable comparison for our method are the following. First, our method and the QP-based {CP controller} are based on the same CP dynamics of the LIPFM and integrate ankle, hip, and stepping strategies that include step time optimization. Second, the QP-based {CP controller \cite{jeong2019robust}} is considered state-of-the-art research in this field. 

The QP formulation proposed by \cite{jeong2019robust} is defined as follows:
\begin{align}
\label{eq/jeong's method Costfunction}
& \underset{\mathbf{f}^{}, \gamma, \bm{\tau}, \mathbf{b}}{\text{min}} & & \mathbf{w}_{f}\| \mathbf{f}_{err}\|^{2} + w_{\gamma}(\gamma^{}_{err})^{2} + \mathbf{w}_{{\tau}}\|\bm{\tau}+\mathbf{K}_{d}\mathbf{h}\|^{2} \nonumber\\ 
& & +&\mathbf{w}_{b}\|\mathbf{b}_{err}\|^{2}  \\ \label{eq/jeong's method constraint}
& \text{s. t.} & & \mathbf{f}_{err} + \mathbf{b}_{err} - \bm{\xi}^{ref}e^{-\omega t}\gamma_{err} -(1-e^{-\omega t}\gamma^{'}){\frac{\bm{\tau}_{err}}{mg}}\nonumber\\ 
& & &=(\bm{\xi}_{err}-\Delta\mathbf{z}_{})e^{-\omega t}\gamma_{err}  + \Delta\mathbf{z}_{}  
\end{align}
where the subscript $err$ signifies the difference between the optimization variables and the pre-designed nominal variables. The delta ZMP $\Delta \mathbf{z}=[\Delta z_x\,\Delta z_y$] denotes the difference between the desired ZMP for controlling the CP and the pre-designed reference ZMP. The cost function (\ref{eq/jeong's method Costfunction}) comprises the step position error term, step time error term, damping term of the CAM, and CP offset error term. The equality constraint in (\ref{eq/jeong's method constraint}) is derived from the error dynamics of the CP end-of-step dynamics in LIPFM. Detailed derivation and explanations for each equation are provided in \cite{jeong2019robust}. 

\begin{figure}[t]
      \centering
      \includegraphics[width=1.0\linewidth]{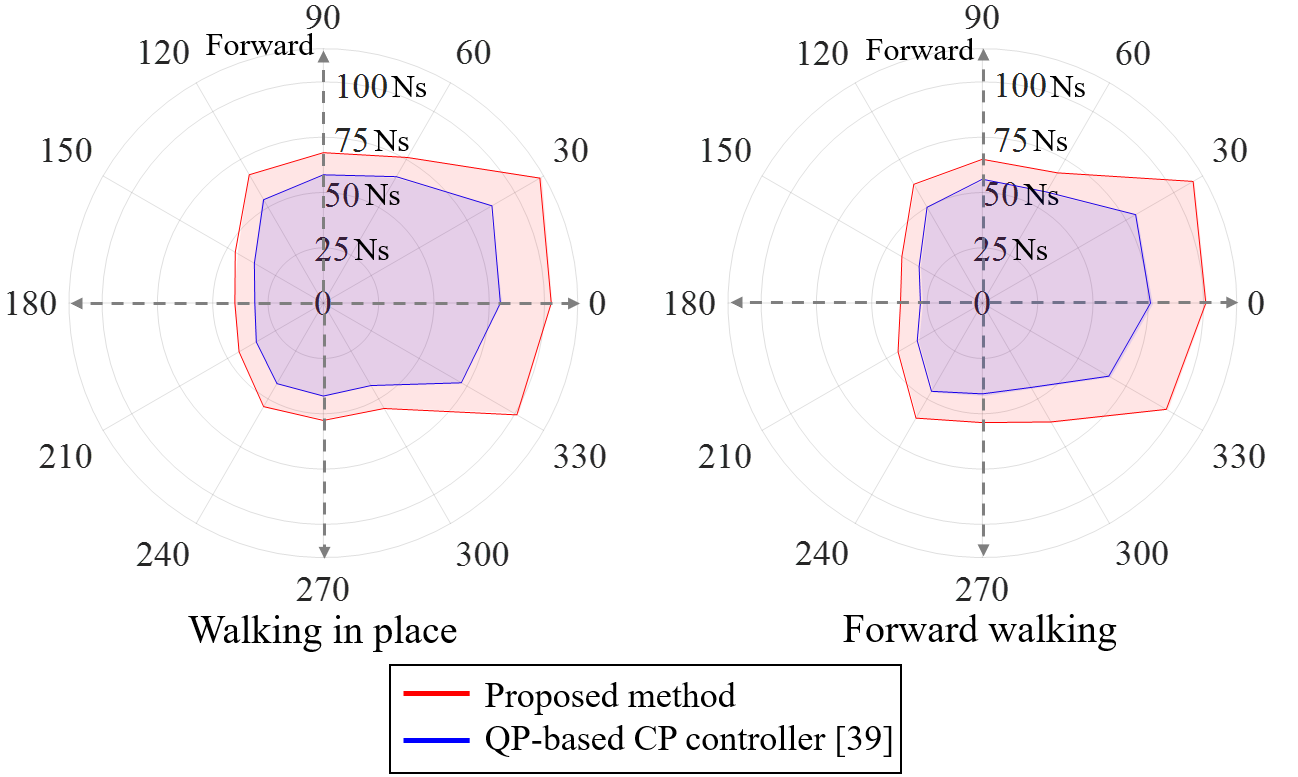} 
      \caption{{Comparisons of the maximum endurable impulse during walking between the proposed method and QP-based CP controller [39].}} 
      \label{figure/Simulation4_DP}
\end{figure}
The QP formulation in (\ref{eq/jeong's method Costfunction}) and (\ref{eq/jeong's method constraint}) addresses the ankle strategy through delta ZMP $\Delta \mathbf{z}\in\mathbb{R}^{2}$, the hip strategy through centroidal moment $\bm{\tau}\in\mathbb{R}^{2}$, and the stepping strategy through footstep position $\mathbf{f}\in\mathbb{R}^{2}$ and step time $\gamma$.
However, due to the variable coupling between $\Delta \mathbf{z}$ and $\gamma$, as well as between $\gamma$ and $\bm{\tau}$, the original equation of (\ref{eq/jeong's method constraint}) resulted in a non-linear constraint. To address this issue in \cite{jeong2019robust}, $\Delta \mathbf{z}$ is considered as a constant using the pre-computed value from the {instantaneous CP end-of-step controller} proposed in \cite{englsberger2011bipedal} without optimization. Furthermore, to avoid variable coupling between $\bm{\tau}$ and $\gamma$, a constant $\gamma'$ was used, which corresponds to the $\gamma$ from the previous control cycle. In contrast, our approach optimizes ZMP, centroidal moment, and footstep position simultaneously using CP--MPC, and the step time is separately optimized in the stepping controller without any constant assumptions.
 \begin{figure*}[t]
      \centering  
      \includegraphics[width=0.98\linewidth]{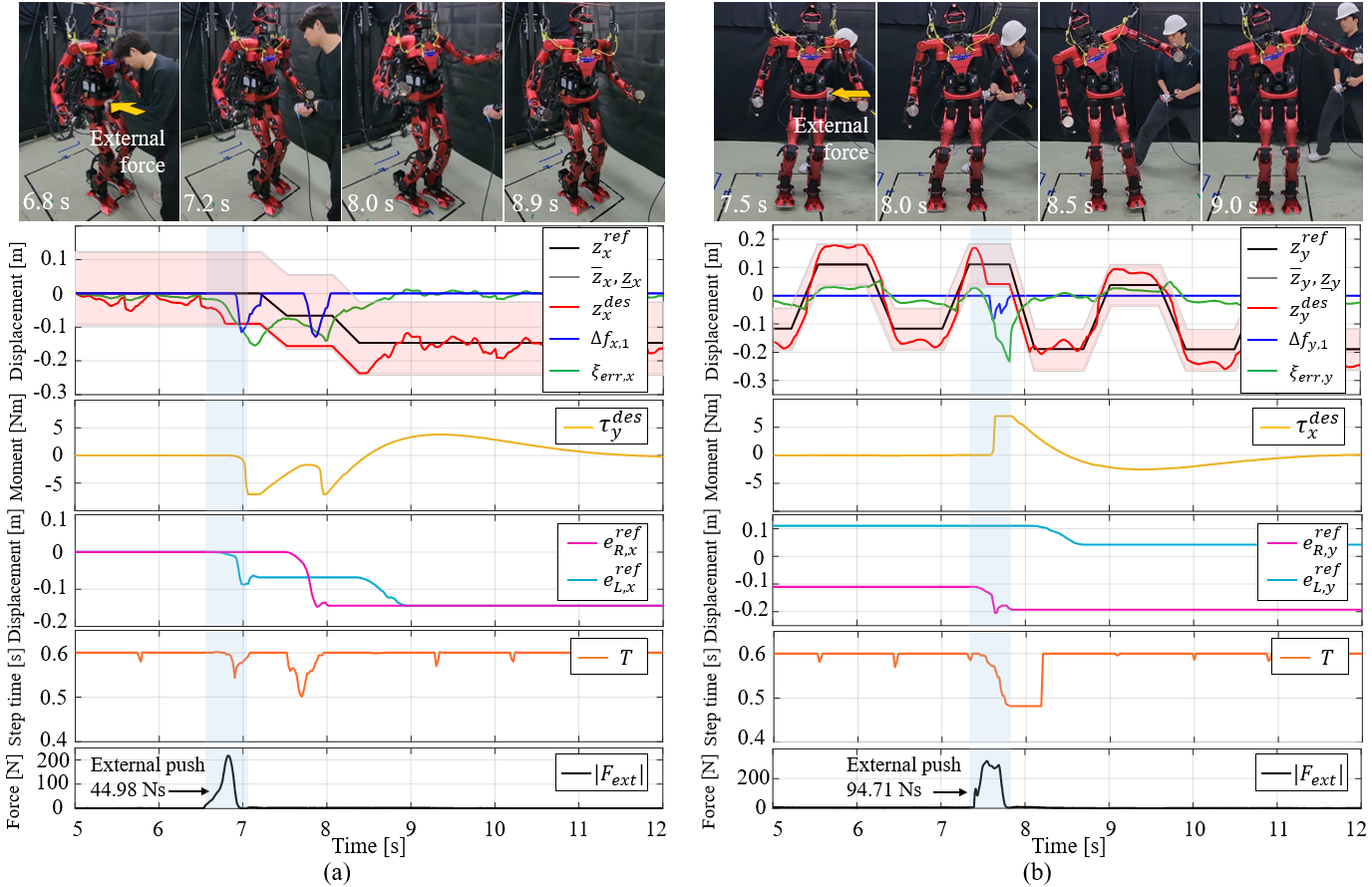} 
      \caption{{Experimental results are presented for the robot's response to external forces applied along both the negative x- and y-directions, with snapshots and output data shown in (a) and (b) respectively.} }
      \label{figure/Experiment_1_Snapshot_Graph}
\end{figure*}

Simulations were conducted to compare the robustness of the two algorithms against external forces. The external forces were applied to the robot during walking in place and forward walking, and the maximum impulse that the robot could withstand was analyzed using DP. The weighting parameter in (\ref{eq/jeong's method Costfunction}) was the same as that used in \cite{jeong2019robust}. Fig. \ref{figure/Simulation4_DP} shows the DP obtained when using both methods. Overall, our method exhibited more robust balancing performance, especially at 0, 30, and 330 degrees, compared to the {QP-based CP controller}. The overall difference in balancing performance may first be attributed to the ankle strategy. 
In \cite{jeong2019robust}, the desired ZMP for instantaneous CP control is calculated as follows,
\begin{align}
\label{eq/jeong_ankle}
\mathbf{z}^{des} &= \mathbf{z}^{ref} + \Delta\mathbf{z} = \mathbf{z}^{ref} - \frac{e^{\omega(T-t)}}{1-e^{\omega(T-t)}}\bm{\xi}_{err}.
\end{align}
However, in (\ref{eq/jeong_ankle}), since the CP control feedback gain is determined based on the remaining step time $(T-t)$, the CP control performance cannot be increased as needed. Next, as analyzed in Section \ref{Section/Results/Simulation/extforce}, stepping control was the most helpful in withstanding external forces at 0, 30, and 330 degrees. 
{The QP-based CP controller \cite{jeong2019robust} considers only the robot's CP for the current step duration, performing instantaneous control to reduce the current CP error. In contrast, our CP--MPC not only considers the current state but also extends its consideration to future constraints and the future state of the robot. This enables proactive stepping control against disturbances by anticipating future CP errors. }
Therefore, the response to disturbances through stepping control was slower compared to our method.
The start time of the stepping control for each method was measured 10 times when the robot was subjected to the maximum external impulse that our method could withstand at 0, 30, and 330 degrees (in Fig. \ref{figure/Simulation4_DP}). Our method, on average, outpaced the {QP-based CP controller} by 0.078 s, 0.086 s, and 0.084 s for each direction, respectively. 
Considering the SSP time (0.6 s), this means that stepping control in our method was performed 13$\%$, 14.3$\%$, and 14$\%$ faster than the {QP-based CP controller}, respectively. 
In conclusion, when using our method compared to the {QP-based CP controller} \cite{jeong2019robust}, the average maximum external impulses that the robot could withstand increased by {27.3 $\%$ for walking in place and 31.8 $\%$ for forward walking.} 
\begin{figure*}[t]
      \centering 
      \includegraphics[width=0.98\linewidth]{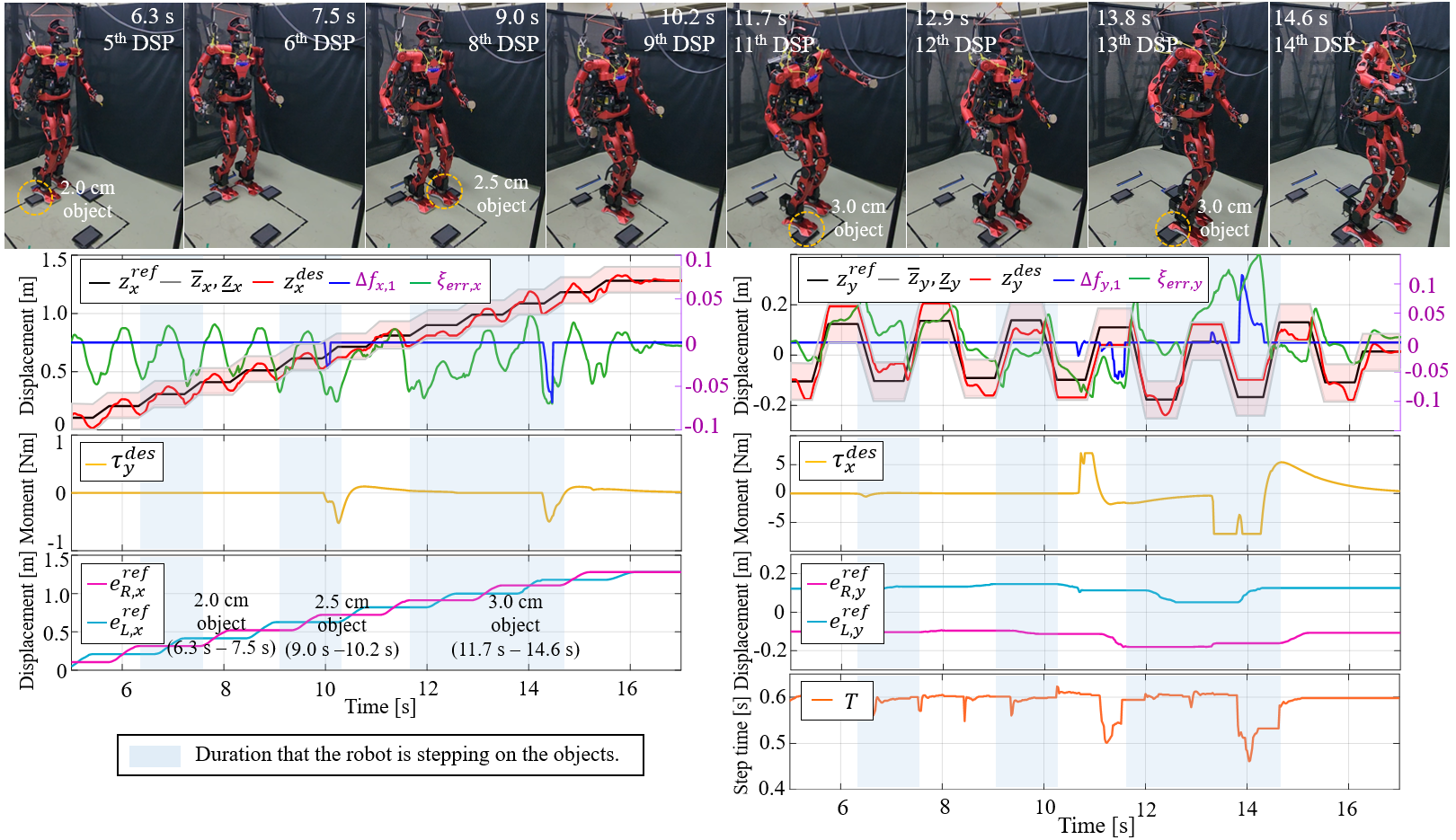} 
      \caption{{Experimental results are presented for the robot's forward walking while overcoming three unexpected objects, with snapshots and output data provided.} }
      \label{figure/Experiment_2_Snapshot_Graph}
\end{figure*}
\subsection{{Real Robot Experiments to Validate Robustness of the Proposed Method under External Forces}}
\label{Section/Results/Experiment}

Real robot experiments were conducted where the robot was subjected to external forces in both x- and y-directions, replicating the scenarios in the simulation presented in Section \ref{Section/Results/Simulation/extforce}. {The external force was applied directly to the robot by a human using a tool equipped with an F/T sensor to measure the timing and magnitude of the force.}
The step duration includes an SSP of 0.6 s and a DSP of 0.3 s which is consistent with the simulation setup. The control parameters used in both the simulation and experiment are identical, except for the constraints of the centroidal moment (as shown in Table. \ref{table/Parameters}). 
The difference in centroidal moment constraints results from conservative upper body motion limits in real robot experiments, applied to prevent collisions with equipment attached to the robot.
Consequently, this implies that if the low-level control of the real robot is already adjusted to suit the hardware, there is no need for tuning of control parameters during sim-to-real implementation.

{In the first experiment, the robot was subjected to an external push of 44.98 Ns in the negative x-direction while in the right foot support. The experimental snapshots and the corresponding data are shown in Fig. \ref{figure/Experiment_1_Snapshot_Graph}(a). 
The snapshot illustrates the robot maintaining balance against external forces. As in Section \ref{Section/Results/Simulation/extforce}, the robot successfully overcame the external force using a combination of ZMP control, CAM control, and stepping control. At approximately {6.5 s}, when the external force is applied to the robot, the CP error $\xi_{err,x}$ (green line) increases in the negative x-direction, and the desired ZMP $z^{des}_x$ (red line) is rapidly generated in the direction that reduces the CP error in the CP--MPC. The desired ZMP is then constrained by the ZMP constraint of -9 cm, and the CP--MPC generates a footstep adjustment $\Delta f_{x,1}$ (blue line) with a peak value of approximately {-11.5 cm to relax the ZMP constraint at approximately 7.0 s}. Additionally, the desired centroidal moment $\tau^{des}_y$ (yellow line) is generated up to -7.0 Nm to further reduce the CP error at approximately {7.1 s.} Despite these efforts, the robot was unable to overcome the strong disturbance. As a result, CP--MPC continuously generated footstep adjustment to reduce the CP error while the desired ZMP was generated up to the constraint until {approximately 8.5 s}. In conclusion, a total of two back-step controls (magenta and cyan lines) were performed, involving a movement of {approximately -15 cm}, along with step time optimization $T$ (orange line) and the robot maintained its balance. In contrast, when the QP-based CP controller proposed in \cite{jeong2019robust} was used in the experiment, despite an external push of 38.7 Ns in the negative x-direction, the CP of the robot could not be recovered and the robot lost its balance.}

{In the following experiment, the robot was subjected to an external push of 94.71 Ns in the negative y-direction during left foot support. The snapshots and the corresponding experimental data are presented in Fig. \ref{figure/Experiment_1_Snapshot_Graph}(b). 
{At approximately 7.4 s,} the external force caused significant perturbation to the robot's CP, resulting in an increase in CP error $\xi_{err,y}$ in the negative y-direction. To reduce the CP error, the desired ZMP $z^{des}_y$ was generated up to the ZMP constraint at {approximately 7.6 s.} The CP--MPC then generated a footstep adjustment $\Delta f_{y,1}$ of approximately {-8.6 cm} and a maximum centroidal moment $\tau_x^{des}$ of 7.0 Nm. Additionally, the stepping controller reduced the step time $T$ {from 0.6 s to 0.48 s.} 
Subsequently, the desired ZMP was continuously generated up to the ZMP constraint to reduce the CP error until {approximately 8.1 s. However, centroidal moment and footstep adjustment were reduced after 7.8 s,} and CP control was handled mainly through ZMP control. Finally, the robot maintained balance against external forces. 
On the other hand, when using the QP-based CP controller \cite{jeong2019robust}, the robot could not withstand an external force of approximately 75.43 Ns and lost balance.
Overall, the magnitude of external impulses applied to the robot was similar to those in the simulation, although it was hard to make them identical due to the human factor. }

\subsection{{Real Robot Experiment to Validate Robustness of the Proposed Method on Uneven Terrain}}

To validate the robustness of the proposed method in uneven terrain, a real robot experiment was conducted. In a scenario similar to the simulation in Section \ref{Section/Results/Simulation/uneven}, three objects, each with a length of 15 cm and width of 20 cm, were placed in front of the robot. The distance between the first and second objects was 25 cm, and the distance between the second and third objects was 40 cm. The objects were placed alternately on the left and right sides of the robot's center.
{The thickness of the objects ranged from 2.0 cm to 3.0 cm in increments of 0.5 cm.} The robot walked a distance of 1.3 m with a step length of 0.1 m, and the walking duration consisted of 0.6 s of SSP and 0.3 s of DSP. 

Fig. \ref{figure/Experiment_2_Snapshot_Graph} presents experimental snapshots and the corresponding output data. The snapshots provide a reaction of the robot to each object before and after stepping on it. 
{The first object was primarily overcome using ZMP control, while the second and third objects were addressed through a combination of stepping and CAM control in addition to ZMP control.}

{Based on the data presented in the graph, the balancing process can be explained. During forward walking, the robot stepped onto a 2.0 cm object at the start of the 5th DSP, approximately 6.3 s. However, the robot’s CP was not significantly perturbed, allowing it to overcome the obstacle primarily through ZMP control until approximately 7.5 s.}


{At the start of the 8th DSP, approximately 9.0 s, the robot stepped at an angle on a 2.5 cm object, resulting in a significant increase in CP error in both the x- and y-directions until the 11th SSP. To reduce this error, CP--MPC not only generates the ZMP up to the ZMP constraints in both x- and y-directions but also generates footstep adjustments $\Delta f_{x,y,1}$ and the desired centroidal moment $\tau_{x,y}^{des}$. The stepping controller then adjusts the swing foot position $e^{ref}_{L,R}$ based on these footstep adjustments $\Delta f_{x,y,1}$ and the step time $T$. Unlike the disturbance from the initial object, all three balance strategies are actively employed.}

{Immediately after the balance control performed until the 11th SSP, from the start of the 11th DSP at 11.7 s to the end of the 14th DSP at 14.6 s, the robot continuously stepped obliquely on a 3.0 cm object in the y-direction. This caused a significantly larger disturbance in the y-direction compared to the x-direction. In response, the robot actively employed the three balance strategies to control the y-direction CP error, ultimately enabling it to maintain balance.}

Similar to the results in Section \ref{Section/Results/Simulation/uneven}, where the robot primarily overcame disturbances in the x-direction using ZMP control with a long support foot length, in the y-direction, both stepping and CAM control were actively utilized. When using the {QP-based CP controller} \cite{jeong2019robust}, however, the robot could not withstand the disturbance from a 3.0 cm object and fell over.


\section{{Discussion}}
\label{Section/Discussion}
\subsection{{Hierarchical Structure of CP--MPC and Stepping Controller}} 
\label{Section/Discussion/Hierarchical}
\subsubsection{Optimality}\label{Section/Discussion/Hierarchical/Optimality}
{In the proposed walking framework, the CP--MPC and the stepping controller are structured hierarchically. 
The CP--MPC is not re-optimized within the same control cycle based on adjusted step time $T$ and footstep position $\mathbf{f}$ from the stepping controller. In order to analyze the impact of the adjusted step time and footstep position on the optimality of CP--MPC, we re-optimized CP--MPC by incorporating the adjusted footstep $\mathbf{f}$ and step time $T$. We then analyzed the cost values (\ref{eq/CP-MPC cost function}) of CP--MPC during this re-optimization process to examine the impact of these adjustments on its optimality.}

\begin{figure}[t]
      \centering
      \includegraphics[width=0.98\linewidth]{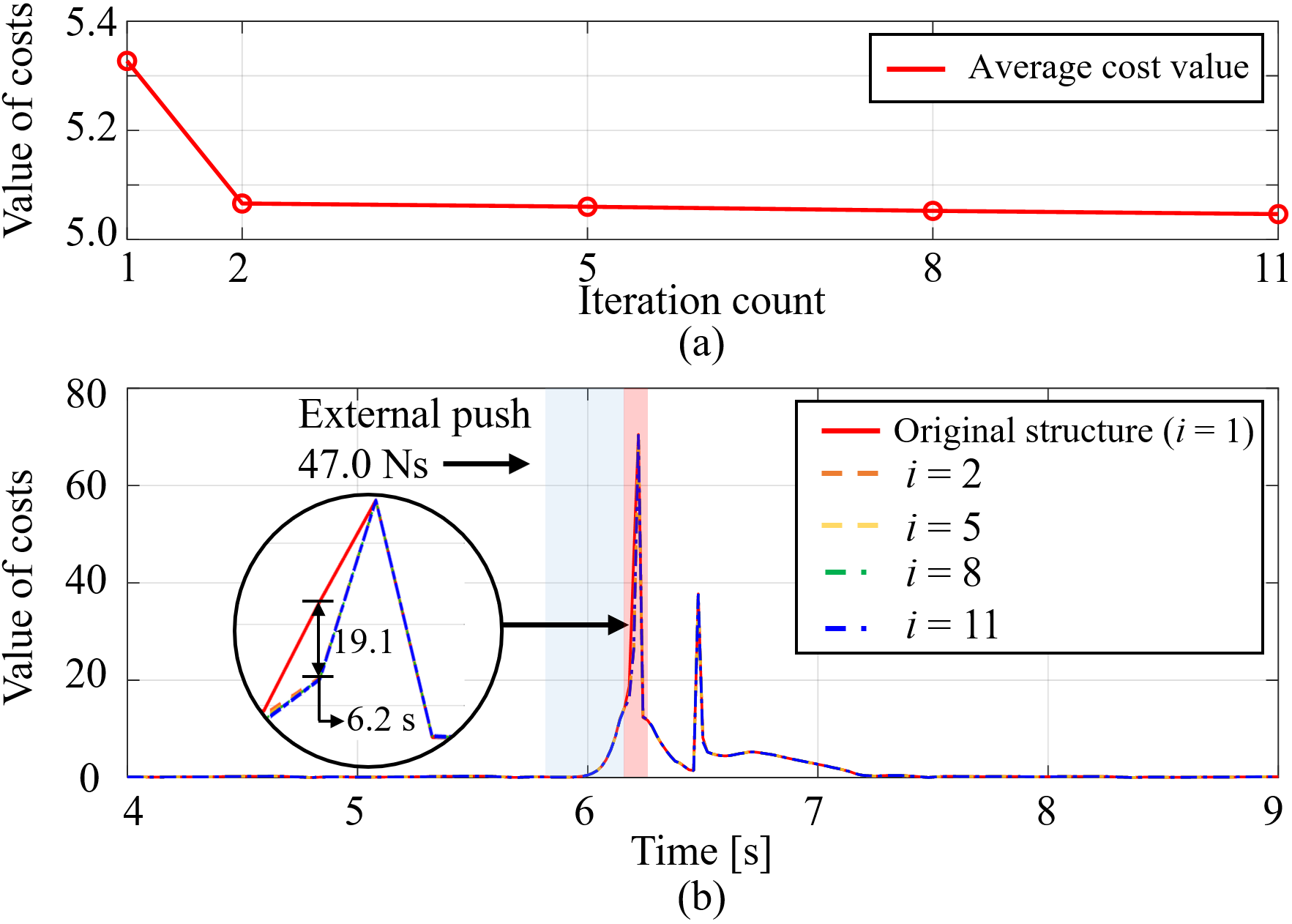} 
      \caption{{Value of costs according to iterative optimization of CP--MPC and the stepping controller}} 
      \label{figure/Simulation_8_rev_cost}
\end{figure}

{Fig. \ref{figure/Simulation_8_rev_cost} depicts the iteration--cost graph in a scenario where a robot experiences disturbances from the front to the back during forward walking. Fig. \ref{figure/Simulation_8_rev_cost}(a) displays the values of the average cost (\ref{eq/CP-MPC cost function}) over the iteration count during the period from 5.9 s, right after the disturbance was applied to the robot, to approximately 7.4 s when the robot stabilizes (two walking steps). Fig. \ref{figure/Simulation_8_rev_cost}(b) illustrates the value of the cost over time, where $i$ denotes the iteration count.} 

{In Fig. \ref{figure/Simulation_8_rev_cost}(a), a decreasing trend in the average cost is evident with iterative re-optimization compared to the original structure ($i=1$). After two iterations ($i=2$), the average cost is reduced by 5.1 $\%$, indicating a noticeable reduction in the second iteration. However, subsequent iterations show diminishing returns in terms of cost reduction. Furthermore, in Fig. \ref{figure/Simulation_8_rev_cost}(b), a difference in cost values between the original structure and $i\geq 2$ is observed at 6.2 s with a magnitude of around 19.1.
This implies that changes in the step time and footstep position from the stepping controller lead to variations in the cost of CP--MPC and affect the optimality of CP--MPC. Therefore, re-optimizing CP--MPC twice by reflecting the adjusted step time and footstep position yields values closer to the optimum for CP--MPC.}

{However, more than two iterations incurs a significant computational load, making it difficult to maintain a 50 Hz operation with our 1.5 s horizon setup. To achieve a two-time iteration, it is necessary to adjust the CP--MPC's operation to 40 Hz and reduce the MPC horizon to 1.0 s.}

{We compared the robustness against disturbances between the two-iteration case and the original framework. 
The robot exhibited a 21.6 $\%$ and 23.2 $\%$ reduction in its maximum external force tolerance during in-place and forward walking, respectively, compared to the original control structure.}
{Consequently, the shortened horizon and reduced control frequency negatively impacted the robot's robustness during walking. From this perspective, as suggested in \cite{scianca2020mpc}, implementing an intrinsically stable MPC that is less dependent on horizon length could be one of the promising ideas for improving the overall performance of MPC.}

\subsubsection{Adavantages and limitations}\label{Section/Discussion/Hierarchical/Advantages}
{The proposed hierarchical structure has the advantage of handling step time adjustment as a linear optimization problem, avoiding the nonlinearity associated with considering step time as an MPC variable. Previous studies have either utilized Sequential Quadratic Programming (SQP) for nonlinear MPC calculations \cite{aftab2012ankle, choe2023seamless} or linearized nonlinear constraints by defining specific bounds for the step time variable \cite{bohorquez2017adaptive}. However, optimization of step time within the MPC, as reported in \cite{khadiv2020walking}, may increase computation time and risk getting stuck in undesired local minima. The increased computational load may compromise robustness against disturbances by MPC optimizations with short horizon length and low control frequencies. In contrast, our approach defines the step time optimization as a convex problem, making it suitable for real-time applications.}

{However, unlike conventional NMPC-based methods \cite{aftab2012ankle, kryczka2015online, bohorquez2017adaptive, choe2023seamless}, our framework optimizes step time without explicitly considering future states and constraints, unlike MPC. Additionally, it may have a limitation where the CP--MPC solution may not be entirely optimal when the step time undergoes significant modifications. Despite this limitation, our framework can enhance robustness against disturbances by operating CP--MPC and the stepping controller with a higher control frequency and sufficient CP--MPC horizon.}
\subsubsection{Time horizon}
{In \cite{zaytsev2015two}, it is reported that considering a future horizon beyond two walking steps in an MPC framework is unnecessary for humanoid balancing, and balancing can be achieved within mostly two steps when disturbances occur. Based on a 0.9 s step duration, we consider a 1.5 s MPC horizon for CP--MPC and a 0.9 s horizon for the stepping controller. Similar to the horizon length suggested in \cite{zaytsev2015two}, our horizon length considers 1 to 2 walking steps for both controllers. Note that, unlike CP--MPC, the stepping controller considers only a one-step horizon, as it relies on CP end-of-step dynamics and the CP--MPC control input for a single walking step. }

\subsection{{Sequence for Ankle, Hip, and Stepping Strategies in Response to Disturbances}}
\label{Section/Discussion/Sequence}
{
During robot walking, small CP errors caused by foot-ground impacts, model uncertainties, etc., can be reduced through ankle strategy (ZMP control). 
However, larger CP errors from external impacts or uneven terrain may exceed the ankle strategy’s capability, requiring additional use of hip (CAM control) and stepping strategies (stepping control).
CP--MPC guides optimal behavior by leveraging the weights of the regulation terms in (\ref{eq/CP-MPC cost function}) ($\mathbf{w}_{\tau}, \mathbf{w}_{F}$), determining which strategy to employ more actively.}

{
In the hip strategy, after balance recovery motions, the robot halts and returns to its initial posture, requiring time to reuse.
Moreover, excessive use of the hip strategy for small CP errors can lead to unnatural motions. Therefore, we employ a variable weighting approach to suppress the use of the hip strategy for small CP errors and allow its flexibility for large CP errors to facilitate balance recovery. 
As a result, the ankle and stepping strategies are employed first, with the hip strategy used as needed, similar to human behavior \cite{park2004postural, mcilroy1993changes, mansfield2011training}.}

\subsection{{CoM State Estimation}}
\label{Section/Discussion/Com}
\subsubsection{CoM position}
{The position of the CoM is estimated by utilizing global orientation information from the IMU and conducting forward kinematics calculations. } 
\subsubsection{CoM velocity}
{The linear velocity of the pelvis is initially estimated using a complementary filter, integrating the linear acceleration from the IMU and the pelvis's linear velocity that is kinematically calculated using joint angular velocities. The angular velocity of the pelvis is directly measured from the IMU. Subsequently, the velocity of the CoM is estimated by utilizing the spatial velocity of the pelvis, joint angular velocities, and the CoM Jacobian.}


\subsection{{Selection of IK over ID Control}} 
\label{Section/Discussion/IKID}
{Our research group is exploring high-performance whole-body ID control {for humanoids using only proprioceptive sensors, without joint torque sensors,} even in the presence of challenges such as friction and elasticity in high-reduction-ratio actuators. In this study, however, we opted for whole-body IK to develop the walking algorithm in parallel, without the presence of such uncertainty factors.}

\subsection{{Adaptation to Early or Late Foot Contacts}} 
{Early or late foot contacts are managed through the ZMP control in \cite{kajita2010biped}. In cases of early or late foot contact, a significant difference arises between the desired contact wrench for achieving the desired ZMP and the measured contact wrench. The ZMP controller adjusts the ankle orientations and the leg length difference between both legs to track the desired contact wrench using admittance control, enabling the robot to maintain balance amidst unexpected early or late foot contacts.}  

\subsection{{Limitations of balancing performance}} 
{From the external force experiments, we found that when addressing disturbances directed toward the stance foot, the kinematic constraints designed to prevent leg collisions hinder the adjustment of far footstep placements, degrading balancing performance. In such cases, strategies like rapid stepping or cross-leg movements could be beneficial.}

{In the uneven terrain experiments, our system, which employs position control with high gear ratio actuators, struggled to quickly respond to early contact with unexpected objects. As a result, unless the experimental walking duration is further extended, the success rate for overcoming objects over 3.5 cm diminishes. We plan to address this issue by researching whole-body ID control, as mentioned in Section \ref{Section/Discussion/IKID}.}

\section{Conclusion}
\label{Section/Conclusion}
This paper presents a novel balance control framework for humanoid robots to achieve robust balancing performance. The framework employs a linear MPC to drive three balance strategies, namely ankle, hip, and stepping strategies, for CP tracking control, effectively calculating ZMP, CAM, and footstep positions. Additionally, a variable weighting method adjusts MPC weighting parameters over the time horizon for CAM damping control, reducing the influence of damping action on CAM and improving balancing performance. 

Next, a hierarchical structure combining CP--MPC and a stepping controller is proposed, enabling step time optimization. Specifically, the stepping controller adjusts the step time while closely following the footstep adjustments planned by CP--MPC to control the CP offset. The parameter selection method in the stepping controller is validated against methods from previous studies \cite{khadiv2016step, khadiv2020walking, kim2023foot}.

Finally, the balancing performance of the proposed method has been validated through simulations and real robot experiments under various scenarios with disturbances. The robot can traverse uneven terrain and overcome external forces.  

{In the current walking framework, LIPFM dynamics are defined under the assumption of flat terrain, with no adjustments made for ground inclination affecting LIPFM dynamics, foot, or CoM trajectories. Our future work will adapt these dynamics and trajectories for sloped or uneven terrain. 
Additionally, to enhance generality, we plan to implement a state machine for step generation in response to disturbances, even after a predefined walking sequence or while standing still.}

{\appendix[A. Derivation of CMP Parameter $\mathbf{P}_{ssp}$] \label{section/appendix}
{According to the definition of (\ref{eq/reculsive dynamics}), the predicted CP end-of-step within the MPC can be expressed as follows,
\begin{equation} \label{eq/CPeos_CPMPC}
    \resizebox{0.89\linewidth}{!}{%
        $
        {\xi}_{T_k,x} = {A}^{T_k-k}\xi_{x,k} + 
        \begin{bmatrix}
        {A}^{T_k-k-1}\mathbf{B} \cdots A^{0}\mathbf{B}
        \end{bmatrix}
        \begin{bmatrix}
        z_{x,k} \\ \tau_{y,k} \\ \vdots \\ z_{x,T_k-1} \\ \tau_{y,T_k-1} 
        \end{bmatrix}
        $
    }
\end{equation}
where $T_k$ represents the time step at which SSP ends within the MPC horizon. To incorporate the predicted CP end-of-step (\ref{eq/CPeos_CPMPC}) into the equality constraint (\ref{eq/stepping_constraint}) of the stepping controller, it is necessary to represent it in the discretized form of the CP end-of-step dynamics (\ref{eq/stepping_cpeos}) as follows,
\begin{align}
\label{appendix/CPeos}
    {\xi}_{T_k,x} &= {A}^{T_k-k}(\xi_{x,k}-{p}_{ssp,x,k})+{p}_{ssp,x,k} \\ \nonumber
    &= {A}^{T_k-k}\xi_{x,k}+(1-{A}^{T_k-k})\,{p}_{ssp,x,k}. 
\end{align} 
Here, $A=e^{\omega T_s}$, and $p_{ssp,x,k}$ denotes the CMP parameter in the x-direction. 
Therefore, to make (25) identical to (24), the CMP parameter ${p}_{ssp,x}$ is defined as follows:
\begin{align}
    \resizebox{0.89\linewidth}{!}{  
        ${p}_{ssp,x} = \frac{1}{1-{A}^{T_k-k}} 
        \begin{bmatrix}
        {A}^{T_k-k-1}\mathbf{B} \cdots {A}^{0}\mathbf{B}
        \end{bmatrix}
        \begin{bmatrix}
        z_{x,k} \\ \tau_{y,k} \\ \vdots \\ z_{x,T_k-1} \\ \tau_{y,T_k-1}
        \end{bmatrix} $
    }
\end{align} 
}

 
%

\bibliographystyle{BIB/IEEEtran}
\bibliography{BIB/Manuscript}

\begin{thebibliography}{10}
\providecommand{\url}[1]{#1}
\csname url@rmstyle\endcsname
\providecommand{\newblock}{\relax}
\providecommand{\bibinfo}[2]{#2}
\providecommand\BIBentrySTDinterwordspacing{\spaceskip=0pt\relax}
\providecommand\BIBentryALTinterwordstretchfactor{4}
\providecommand\BIBentryALTinterwordspacing{\spaceskip=\fontdimen2\font plus
\BIBentryALTinterwordstretchfactor\fontdimen3\font minus \fontdimen4\font\relax}
\providecommand\BIBforeignlanguage[2]{{%
\expandafter\ifx\csname l@#1\endcsname\relax
\typeout{** WARNING: IEEEtran.bst: No hyphenation pattern has been}%
\typeout{** loaded for the language `#1'. Using the pattern for}%
\typeout{** the default language instead.}%
\else
\language=\csname l@#1\endcsname
\fi
#2}}

\bibitem{nashner1985organization}
L.~M. Nashner and G.~McCollum, ``The organization of human postural movements: a formal basis and experimental synthesis,'' \emph{Behavioral and brain sciences}, vol.~8, no.~1, pp. 135--150, 1985.

\bibitem{winter1995human}
D.~A. Winter, ``Human balance and posture control during standing and walking,'' \emph{Gait \& posture}, vol.~3, no.~4, pp. 193--214, 1995.

\bibitem{maki1997role}
B.~E. Maki and W.~E. McIlroy, ``The role of limb movements in maintaining upright stance: the “change-in-support” strategy,'' \emph{Physical therapy}, vol.~77, no.~5, pp. 488--507, 1997.

\bibitem{barin1989evaluation}
K.~Barin, ``Evaluation of a generalized model of human postural dynamics and control in the sagittal plane,'' \emph{Biological cybernetics}, vol.~61, no.~1, pp. 37--50, 1989.

\bibitem{kuo1993human}
A.~D. Kuo and F.~E. Zajac, ``Human standing posture: multi-joint movement strategies based on biomechanical constraints,'' \emph{Progress in brain research}, vol.~97, pp. 349--358, 1993.

\bibitem{park2004postural}
S.~Park, F.~B. Horak, and A.~D. Kuo, ``Postural feedback responses scale with biomechanical constraints in human standing,'' \emph{Experimental brain research}, vol. 154, no.~4, pp. 417--427, 2004.

\bibitem{kajita20013d}
S.~Kajita, F.~Kanehiro, K.~Kaneko, K.~Yokoi, and H.~Hirukawa, ``The 3d linear inverted pendulum mode: A simple modeling for a biped walking pattern generation,'' in \emph{Proceedings 2001 IEEE/RSJ International Conference on Intelligent Robots and Systems. Expanding the Societal Role of Robotics in the the Next Millennium (Cat. No. 01CH37180)}, vol.~1.\hskip 1em plus 0.5em minus 0.4em\relax IEEE, 2001, pp. 239--246.

\bibitem{kajita2003biped}
S.~Kajita, F.~Kanehiro, K.~Kaneko, K.~Fujiwara, K.~Yokoi, and H.~Hirukawa, ``Biped walking pattern generation by a simple three-dimensional inverted pendulum model,'' \emph{Advanced Robotics}, vol.~17, no.~2, pp. 131--147, 2003.

\bibitem{vukobratovic2004zero}
M.~Vukobratovi{\'c} and B.~Borovac, ``Zero-moment point—thirty five years of its life,'' \emph{International journal of humanoid robotics}, vol.~1, no.~01, pp. 157--173, 2004.

\bibitem{popovic2005ground}
M.~B. Popovic, A.~Goswami, and H.~Herr, ``Ground reference points in legged locomotion: Definitions, biological trajectories and control implications,'' \emph{The international journal of robotics research}, vol.~24, no.~12, pp. 1013--1032, 2005.

\bibitem{kajita2003preview}
S.~Kajita, F.~Kanehiro, K.~Kaneko, K.~Fujiwara, K.~Harada, K.~Yokoi, and H.~Hirukawa, ``Biped walking pattern generation by using preview control of zero-moment point,'' in \emph{2003 IEEE international conference on robotics and automation (Cat. No. 03CH37422)}, vol.~2.\hskip 1em plus 0.5em minus 0.4em\relax IEEE, 2003, pp. 1620--1626.

\bibitem{choi2006walking}
Y.~Choi, D.~Kim, and B.-J. You, ``On the walking control for humanoid robot based on the kinematic resolution of com jacobian with embedded motion,'' in \emph{Proceedings 2006 IEEE International Conference on Robotics and Automation, 2006. ICRA 2006.}\hskip 1em plus 0.5em minus 0.4em\relax IEEE, 2006, pp. 2655--2660.

\bibitem{kim2006experimental}
J.-Y. Kim, I.-W. Park, and J.-H. Oh, ``Experimental realization of dynamic walking of the biped humanoid robot khr-2 using zero moment point feedback and inertial measurement,'' \emph{Advanced Robotics}, vol.~20, no.~6, pp. 707--736, 2006.

\bibitem{kajita2010biped}
S.~Kajita, M.~Morisawa, K.~Miura, S.~Nakaoka, K.~Harada, K.~Kaneko, F.~Kanehiro, and K.~Yokoi, ``Biped walking stabilization based on linear inverted pendulum tracking,'' in \emph{2010 IEEE/RSJ International Conference on Intelligent Robots and Systems}.\hskip 1em plus 0.5em minus 0.4em\relax IEEE, 2010, pp. 4489--4496.

\bibitem{joe2019robust}
H.-M. Joe and J.-H. Oh, ``A robust balance-control framework for the terrain-blind bipedal walking of a humanoid robot on unknown and uneven terrain,'' \emph{Sensors}, vol.~19, no.~19, p. 4194, 2019.

\bibitem{komura2005feedback}
T.~Komura, H.~Leung, S.~Kudoh, and J.~Kuffner, ``A feedback controller for biped humanoids that can counteract large perturbations during gait,'' in \emph{Proceedings of the 2005 IEEE International Conference on Robotics and Automation}.\hskip 1em plus 0.5em minus 0.4em\relax IEEE, 2005, pp. 1989--1995.

\bibitem{komura2005simulating}
T.~Komura, A.~Nagano, H.~Leung, and Y.~Shinagawa, ``Simulating pathological gait using the enhanced linear inverted pendulum model,'' \emph{IEEE Transactions on biomedical engineering}, vol.~52, no.~9, pp. 1502--1513, 2005.

\bibitem{pratt2006capture}
J.~Pratt, J.~Carff, S.~Drakunov, and A.~Goswami, ``Capture point: A step toward humanoid push recovery,'' in \emph{2006 6th IEEE-RAS international conference on humanoid robots}.\hskip 1em plus 0.5em minus 0.4em\relax IEEE, 2006, pp. 200--207.

\bibitem{stephens2007humanoid}
B.~Stephens, ``Humanoid push recovery,'' in \emph{2007 7th IEEE-RAS International Conference on Humanoid Robots}.\hskip 1em plus 0.5em minus 0.4em\relax IEEE, 2007, pp. 589--595.

\bibitem{yi2016whole}
S.-J. Yi, B.-T. Zhang, D.~Hong, and D.~D. Lee, ``Whole-body balancing walk controller for position controlled humanoid robots,'' \emph{International Journal of Humanoid Robotics}, vol.~13, no.~01, p. 1650011, 2016.

\bibitem{wiedebach2016walking}
G.~Wiedebach, S.~Bertrand, T.~Wu, L.~Fiorio, S.~McCrory, R.~Griffin, F.~Nori, and J.~Pratt, ``Walking on partial footholds including line contacts with the humanoid robot atlas,'' in \emph{2016 IEEE-RAS 16th International Conference on Humanoid Robots (Humanoids)}.\hskip 1em plus 0.5em minus 0.4em\relax IEEE, 2016, pp. 1312--1319.

\bibitem{schuller2021online}
R.~Schuller, G.~Mesesan, J.~Englsberger, J.~Lee, and C.~Ott, ``Online centroidal angular momentum reference generation and motion optimization for humanoid push recovery,'' \emph{IEEE Robotics and Automation Letters}, vol.~6, no.~3, pp. 5689--5696, 2021.

\bibitem{kim2022humanoid}
M.-J. Kim, D.~Lim, G.~Park, and J.~Park, ``Humanoid balance control using centroidal angular momentum based on hierarchical quadratic programming,'' in \emph{2022 IEEE/RSJ International Conference on Intelligent Robots and Systems (IROS)}.\hskip 1em plus 0.5em minus 0.4em\relax IEEE, 2022, pp. 6753--6760.

\bibitem{ding2022dynamic}
Y.~Ding, C.~Khazoom, M.~Chignoli, and S.~Kim, ``Dynamic walking with footstep adaptation on the mit humanoid via linear model predictive control,'' \emph{arXiv preprint arXiv:2205.15443}, 2022.

\bibitem{herdt2010online}
A.~Herdt, H.~Diedam, P.-B. Wieber, D.~Dimitrov, K.~Mombaur, and M.~Diehl, ``Online walking motion generation with automatic footstep placement,'' \emph{Advanced Robotics}, vol.~24, no. 5-6, pp. 719--737, 2010.

\bibitem{khadiv2016step}
M.~Khadiv, A.~Herzog, S.~A.~A. Moosavian, and L.~Righetti, ``Step timing adjustment: A step toward generating robust gaits,'' in \emph{2016 IEEE-RAS 16th International Conference on Humanoid Robots (Humanoids)}.\hskip 1em plus 0.5em minus 0.4em\relax IEEE, 2016, pp. 35--42.

\bibitem{jeong2017biped}
H.~Jeong, O.~Sim, H.~Bae, K.~Lee, J.~Oh, and J.-H. Oh, ``Biped walking stabilization based on foot placement control using capture point feedback,'' in \emph{2017 IEEE/RSJ International Conference on Intelligent Robots and Systems (IROS)}.\hskip 1em plus 0.5em minus 0.4em\relax IEEE, 2017, pp. 5263--5269.

\bibitem{joe2018balance}
H.-M. Joe and J.-H. Oh, ``Balance recovery through model predictive control based on capture point dynamics for biped walking robot,'' \emph{Robotics and Autonomous Systems}, vol. 105, pp. 1--10, 2018.

\bibitem{jeong2019robust1}
H.~Jeong, I.~Lee, O.~Sim, K.~Lee, and J.-H. Oh, ``A robust walking controller optimizing step position and step time that exploit advantages of footed robot,'' \emph{Robotics and Autonomous Systems}, vol. 113, pp. 10--22, 2019.

\bibitem{khadiv2020walking}
M.~Khadiv, A.~Herzog, S.~A.~A. Moosavian, and L.~Righetti, ``Walking control based on step timing adaptation,'' \emph{IEEE Transactions on Robotics}, vol.~36, no.~3, pp. 629--643, 2020.

\bibitem{kim2023foot}
M.-J. Kim, D.~Lim, G.~Park, and J.~Park, ``Foot stepping algorithm of humanoids with double support time adjustment based on capture point control,'' in \emph{2023 IEEE International Conference on Robotics and Automation (ICRA)}.\hskip 1em plus 0.5em minus 0.4em\relax IEEE, 2023, pp. 12\,198--12\,204.

\bibitem{wieber2006trajectory}
P.-B. Wieber, ``Trajectory free linear model predictive control for stable walking in the presence of strong perturbations,'' in \emph{2006 6th IEEE-RAS International Conference on Humanoid Robots}.\hskip 1em plus 0.5em minus 0.4em\relax IEEE, 2006, pp. 137--142.

\bibitem{shafiee2017robust}
M.~Shafiee-Ashtiani, A.~Yousefi-Koma, and M.~Shariat-Panahi, ``Robust bipedal locomotion control based on model predictive control and divergent component of motion,'' in \emph{2017 IEEE International Conference on Robotics and Automation (ICRA)}.\hskip 1em plus 0.5em minus 0.4em\relax IEEE, 2017, pp. 3505--3510.

\bibitem{ding2021versatile}
J.~Ding, S.~Xin, T.~L. Lam, and S.~Vijayakumar, ``Versatile locomotion by integrating ankle, hip, stepping, and height variation strategies,'' in \emph{2021 IEEE International Conference on Robotics and Automation (ICRA)}.\hskip 1em plus 0.5em minus 0.4em\relax IEEE, 2021, pp. 2957--2963.

\bibitem{ding2022robust}
J.~Ding, L.~Han, L.~Ge, Y.~Liu, and J.~Pang, ``Robust locomotion exploiting multiple balance strategies: An observer-based cascaded model predictive control approach,'' \emph{IEEE/ASME Transactions on Mechatronics}, vol.~27, no.~4, pp. 2089--2097, 2022.

\bibitem{romualdi2022online}
G.~Romualdi, S.~Dafarra, G.~L'Erario, I.~Sorrentino, S.~Traversaro, and D.~Pucci, ``Online non-linear centroidal mpc for humanoid robot locomotion with step adjustment,'' in \emph{2022 International Conference on Robotics and Automation (ICRA)}.\hskip 1em plus 0.5em minus 0.4em\relax IEEE, 2022, pp. 10\,412--10\,419.

\bibitem{aftab2012ankle}
Z.~Aftab, T.~Robert, and P.-B. Wieber, ``Ankle, hip and stepping strategies for humanoid balance recovery with a single model predictive control scheme,'' in \emph{2012 12th IEEE-RAS International Conference on Humanoid Robots (Humanoids 2012)}.\hskip 1em plus 0.5em minus 0.4em\relax IEEE, 2012, pp. 159--164.

\bibitem{nazemi2017reactive}
F.~Nazemi, A.~Yousefi-Koma, M.~Khadiv, \emph{et~al.}, ``A reactive and efficient walking pattern generator for robust bipedal locomotion,'' in \emph{2017 5th RSI International Conference on Robotics and Mechatronics (ICRoM)}.\hskip 1em plus 0.5em minus 0.4em\relax IEEE, 2017, pp. 364--369.

\bibitem{jeong2019robust}
H.~Jeong, I.~Lee, J.~Oh, K.~K. Lee, and J.-H. Oh, ``A robust walking controller based on online optimization of ankle, hip, and stepping strategies,'' \emph{IEEE Transactions on Robotics}, vol.~35, no.~6, pp. 1367--1386, 2019.

\bibitem{englsberger2011bipedal}
J.~Englsberger, C.~Ott, M.~A. Roa, A.~Albu-Sch{\"a}ffer, and G.~Hirzinger, ``Bipedal walking control based on capture point dynamics,'' in \emph{2011 IEEE/RSJ International Conference on Intelligent Robots and Systems}.\hskip 1em plus 0.5em minus 0.4em\relax IEEE, 2011, pp. 4420--4427.

\bibitem{hof2008extrapolated}
A.~L. Hof, ``The ‘extrapolated center of mass’ concept suggests a simple control of balance in walking,'' \emph{Human movement science}, vol.~27, no.~1, pp. 112--125, 2008.

\bibitem{englsberger2016combining}
J.~Englsberger, ``Combining reduced dynamics models and whole-body control for agile humanoid locomotion,'' Ph.D. dissertation, Technische Universit{\"a}t M{\"u}nchen, 2016.

\bibitem{morisawa2012balance}
M.~Morisawa, S.~Kajita, F.~Kanehiro, K.~Kaneko, K.~Miura, and K.~Yokoi, ``Balance control based on capture point error compensation for biped walking on uneven terrain,'' in \emph{2012 12th IEEE-RAS International Conference on Humanoid Robots (Humanoids 2012)}.\hskip 1em plus 0.5em minus 0.4em\relax IEEE, 2012, pp. 734--740.

\bibitem{griffin2017walking}
R.~J. Griffin, G.~Wiedebach, S.~Bertrand, A.~Leonessa, and J.~Pratt, ``Walking stabilization using step timing and location adjustment on the humanoid robot, atlas,'' in \emph{2017 IEEE/RSJ International Conference on Intelligent Robots and Systems (IROS)}.\hskip 1em plus 0.5em minus 0.4em\relax IEEE, 2017, pp. 667--673.

\bibitem{park2006contact}
J.~Park and O.~Khatib, ``Contact consistent control framework for humanoid robots,'' in \emph{Proceedings 2006 IEEE International Conference on Robotics and Automation, 2006. ICRA 2006.}\hskip 1em plus 0.5em minus 0.4em\relax IEEE, 2006, pp. 1963--1969.

\bibitem{kanamiya2010ankle}
Y.~Kanamiya, S.~Ota, and D.~Sato, ``Ankle and hip balance control strategies with transitions,'' in \emph{2010 IEEE International Conference on Robotics and Automation}.\hskip 1em plus 0.5em minus 0.4em\relax IEEE, 2010, pp. 3446--3451.

\bibitem{khazoom2022humanoid}
C.~Khazoom and S.~Kim, ``Humanoid arm motion planning for improved disturbance recovery using model hierarchy predictive control,'' in \emph{2022 International Conference on Robotics and Automation (ICRA)}.\hskip 1em plus 0.5em minus 0.4em\relax IEEE, 2022, pp. 6607--6613.

\bibitem{park2021whole}
B.~Park, M.-J. Kim, E.~Sung, J.~Kim, and J.~Park, ``Whole-body walking pattern using pelvis-rotation for long stride and arm swing for yaw angular momentum compensation,'' in \emph{2020 IEEE-RAS 20th International Conference on Humanoid Robots (Humanoids)}.\hskip 1em plus 0.5em minus 0.4em\relax IEEE, 2021, pp. 47--52.

\bibitem{kryczka2015online}
P.~Kryczka, P.~Kormushev, N.~G. Tsagarakis, and D.~G. Caldwell, ``Online regeneration of bipedal walking gait pattern optimizing footstep placement and timing,'' in \emph{2015 IEEE/RSJ International Conference on Intelligent Robots and Systems (IROS)}.\hskip 1em plus 0.5em minus 0.4em\relax IEEE, 2015, pp. 3352--3357.

\bibitem{todorov2012mujoco}
E.~Todorov, T.~Erez, and Y.~Tassa, ``Mujoco: A physics engine for model-based control,'' in \emph{2012 IEEE/RSJ international conference on intelligent robots and systems}.\hskip 1em plus 0.5em minus 0.4em\relax IEEE, 2012, pp. 5026--5033.

\bibitem{ferreau2014qpoases}
H.~J. Ferreau, C.~Kirches, A.~Potschka, H.~G. Bock, and M.~Diehl, ``qpoases: A parametric active-set algorithm for quadratic programming,'' \emph{Mathematical Programming Computation}, vol.~6, pp. 327--363, 2014.

\bibitem{felis2017rbdl}
M.~L. Felis, ``Rbdl: an efficient rigid-body dynamics library using recursive algorithms,'' \emph{Autonomous Robots}, vol.~41, no.~2, pp. 495--511, 2017.

\bibitem{scianca2020mpc}
N.~Scianca, D.~De~Simone, L.~Lanari, and G.~Oriolo, ``Mpc for humanoid gait generation: Stability and feasibility,'' \emph{IEEE Transactions on Robotics}, vol.~36, no.~4, pp. 1171--1188, 2020.

\bibitem{choe2023seamless}
J.~Choe, J.-H. Kim, S.~Hong, J.~Lee, and H.-W. Park, ``Seamless reaction strategy for bipedal locomotion exploiting real-time nonlinear model predictive control,'' \emph{IEEE Robotics and Automation Letters}, pp. 1--8, 2023.

\bibitem{bohorquez2017adaptive}
N.~Boh{\'o}rquez and P.-B. Wieber, ``Adaptive step duration in biped walking: a robust approach to nonlinear constraints,'' in \emph{2017 IEEE-RAS 17th International Conference on Humanoid Robotics (Humanoids)}.\hskip 1em plus 0.5em minus 0.4em\relax IEEE, 2017, pp. 724--729.

\bibitem{zaytsev2015two}
P.~Zaytsev, S.~J. Hasaneini, and A.~Ruina, ``Two steps is enough: No need to plan far ahead for walking balance,'' in \emph{2015 IEEE International Conference on Robotics and Automation (ICRA)}.\hskip 1em plus 0.5em minus 0.4em\relax IEEE, 2015, pp. 6295--6300.

\bibitem{mcilroy1993changes}
W.~McIlroy and B.~Maki, ``Changes in early ‘automatic’postural responses associated with the prior-planning and execution of a compensatory step,'' \emph{Brain research}, vol. 631, no.~2, pp. 203--211, 1993.

\bibitem{mansfield2011training}
A.~Mansfield, E.~L. Inness, J.~Komar, L.~Biasin, K.~Brunton, B.~Lakhani, and W.~E. McIlroy, ``Training rapid stepping responses in an individual with stroke,'' \emph{Physical therapy}, vol.~91, no.~6, pp. 958--969, 2011.

\end{thebibliography}

\vfill

\end{document}